\documentclass[11pt]{article}

\usepackage[final]{acl}

\usepackage{times}
\usepackage{latexsym}
\usepackage{amsmath, amssymb} 

\usepackage[T1]{fontenc}

\usepackage[utf8]{inputenc}

\usepackage{microtype}

\usepackage{inconsolata}
\usepackage{booktabs}
\usepackage{multirow}
\usepackage{xcolor}
\usepackage{colortbl}
\usepackage{subcaption}
\usepackage{ragged2e}
\usepackage{tabularx}

\usepackage{graphicx}
\usepackage{hyperref}

\title{UniCreative: Unifying Long-form Logic and Short-form Sparkle via Reference-Free Reinforcement Learning}

\author{
  {\bf Xiaolong Wei}$^{1}$$^*$,
  {\bf Zerun Zhu}$^{2}$$^*$,
  {\bf Simin Niu}$^{3}$,
  {\bf Xingyu Zhang}$^{2}$,
  {\bf Peiying Yu}$^{2}$,
  {\bf Changxuan Xiao}$^{2}$ \\
  {\bf Yuchen Li}$^{2}$, 
  {\bf Jicheng Yang}$^{2}$, 
  {\bf Zhejun Zhao}$^{2}$$^{\dag}$,
  {\bf Chong Meng}$^{2}$, 
  {\bf Long Xia}$^{2}$, 
  {\bf Daiting Shi}$^{2}$ \\
  $^{1}$Beihang University \ \ 
  $^{2}$Baidu Inc. \\
  $^{3}$Renmin University of China \\
  \texttt{xiaolongwei@buaa.edu.cn, zhaozhejun@baidu.com}
}

\begin{document}
\maketitle

\def\thefootnote{*}\footnotetext{Co-first authors with equal contributions.}
\def\thefootnote{\dag}\footnotetext{Corresponding author}

\begin{abstract}
A fundamental challenge in creative writing lies in reconciling the inherent tension between maintaining global coherence in long-form narratives and preserving local expressiveness in short-form texts.
While long-context generation necessitates explicit macroscopic planning, short-form creativity often demands spontaneous, constraint-free expression.
Existing alignment paradigms, however, typically employ static reward signals and rely heavily on high-quality supervised data, which is costly and difficult to scale.
To address this, we propose \textbf{UniCreative}, a unified reference-free reinforcement learning framework.
We first introduce \textbf{AC-GenRM}, an adaptive constraint-aware reward model that dynamically synthesizes query-specific criteria to provide fine-grained preference judgments.
Leveraging these signals, we propose \textbf{ACPO}, a policy optimization algorithm that aligns models with human preferences across both content quality and structural paradigms without supervised fine-tuning and ground-truth references.
Empirical results demonstrate that AC-GenRM aligns closely with expert evaluations, while ACPO significantly enhances performance across diverse writing tasks.
Crucially, our analysis reveals an emergent meta-cognitive ability: the model learns to autonomously differentiate between tasks requiring rigorous planning and those favoring direct generation, validating the effectiveness of our direct alignment approach. Our code is available at \url{https://github.com/weixiaolong94-hub/UniCreative}.
\end{abstract}

\section{Introduction}

Large Language Models (LLMs) have demonstrated impressive fluency in general-purpose text generation; however, their capabilities in \textit{creative writing} remain fundamentally constrained. A central challenge lies in the inherent tension between maintaining \textit{global coherence} in long-form narratives and preserving \textit{local expressiveness} in short creative texts. Despite the rapid expansion of context windows, recent benchmarks reveal that autoregressive generation still struggles with long-range consistency, often exhibiting topic drift, repetition, and structural degradation in extended outputs such as stories or scripts \cite{liu2024lost,bai2025longbench,wu2025longwriter,zhao2025assessing}. These findings suggest that scaling context length alone is insufficient to resolve the structural limitations of current generation paradigms.

\begin{figure}[t]
    \centering

    \begin{subfigure}[b]{0.49\linewidth} 
        \includegraphics[width=\linewidth]{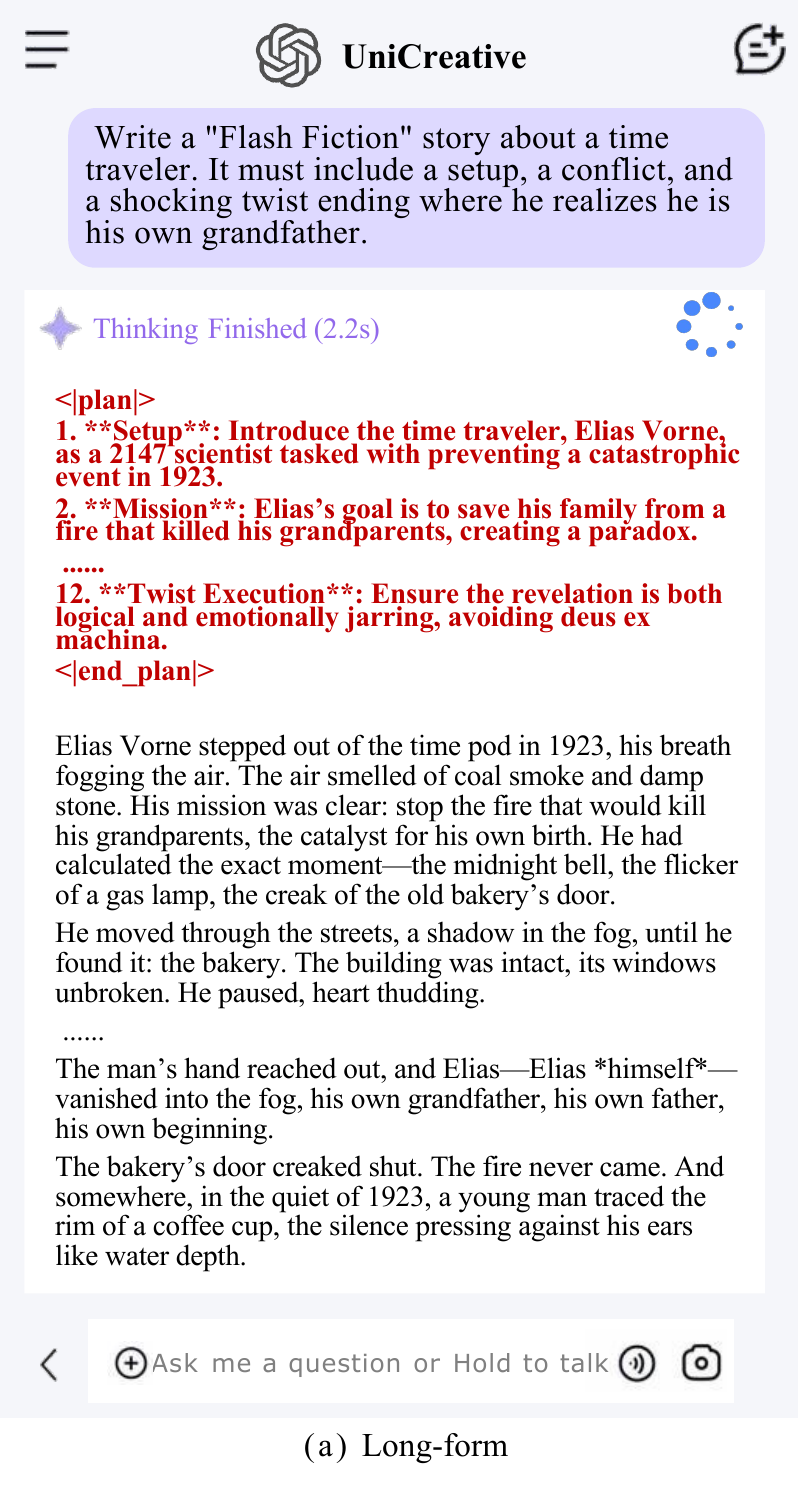}
        \label{fig:sub_long_form}
    \end{subfigure}
    \hfill 
    \begin{subfigure}[b]{0.49\linewidth}
        \includegraphics[width=\linewidth]{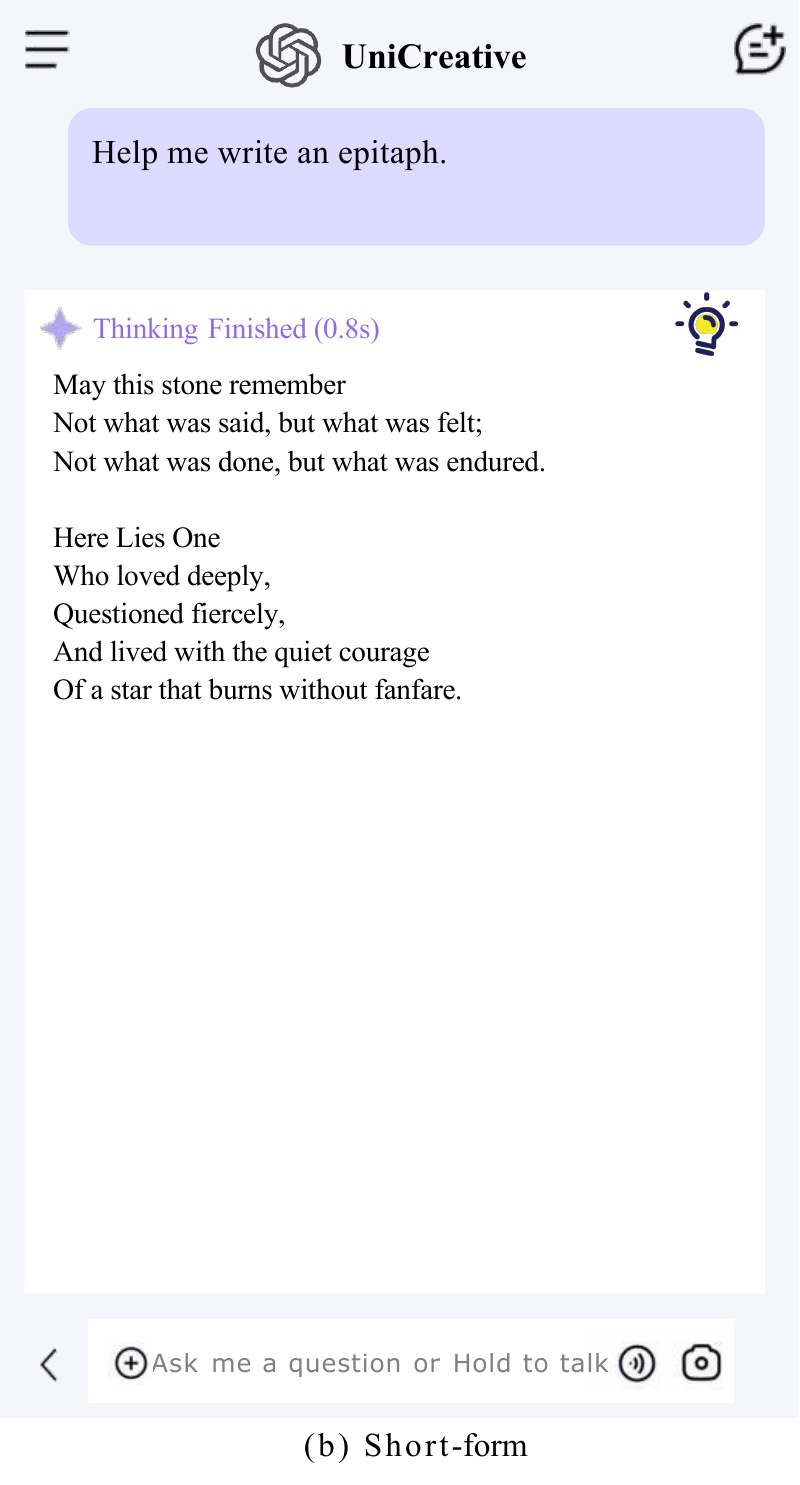}
        \label{fig:sub_short_form}
    \end{subfigure}
    \caption{Examples of UniCreative generations. The long-form task (left) follows a Plan-then-Write procedure, while the short-form task (right) employs direct generation without intermediate planning.}
    \label{fig:long_short_case}
\end{figure}

In stark contrast, short-form creative expression---spanning poetry, slogans, and blessings---necessitates \textit{instantaneous linguistic vibrancy} rather than structural durability. Here, the primary challenge is not logical coherence, but the suppression of statistical mediocrity. Constrained by likelihood maximization, autoregressive models naturally converge toward high-probability, ``safe'' tokens, resulting in homogenized outputs that are fluent yet devoid of the stochastic spark essential for emotional resonance \cite{hou2025creativityprism}. Imposing rigid planning mechanisms on such tasks further exacerbates this issue, leading to a phenomenon of ``over-determination'' where the potential for serendipitous discovery is stifled by premature structural constraints \cite{wu2025writingbench,fein2025litbench,mizrahi2025cooking}. These distinct failure modes—structural collapse in long-form versus creative banality in short-form—highlight that creative writing encompasses heterogeneous regimes that cannot be resolved by a monolithic generation strategy.

A common approach to improving long-form coherence is to introduce explicit planning or hierarchical structure. Early work on hierarchical neural story generation and plan-and-write frameworks demonstrates that global outlines can effectively guide extended narratives and mitigate structural collapse \cite{fan2018hierarchical,yao2019plan}. More recent studies similarly suggest that structured representations and multi-stage generation pipelines can enhance coherence and controllability in complex creative tasks \cite{xiao2025longweave,alper2025conlangcrafter}. However, these same mechanisms can become counterproductive when indiscriminately applied to short creative texts, where excessive structural constraints risk suppressing expressive diversity and creative spontaneity. This contrast underscores that planning should not be treated as a universally optimal solution, but rather as a capability that must be invoked selectively based on the task nature.

Meanwhile, advances in reinforcement learning and preference-based alignment provide new tools for optimizing language models. However, standard paradigms like RLHF or DPO typically rely on high-quality supervised data (SFT) and human-annotated preference pairs, which are costly to collect and difficult to scale for open-ended creative tasks \cite{christiano2017deep,rafailov2023direct}. While recent work has begun to systematically study creativity assessment and modeling in large language models \cite{zhao2025assessing,li2025automated,huang2025causality}, most methods still depend on ground-truth references to compute rewards or stabilize training, limiting their applicability in purely creative domains where no single ``correct'' answer exists. Creative generation systems designed for scientific ideation further highlight the need for reference-free optimization paradigms \cite{sanyal2025spark}.

In this paper, we propose \textbf{UniCreative}, a unified creative writing framework that treats planning as a dynamically callable computational resource rather than a fixed prerequisite. UniCreative enables models to adaptively switch between a \textbf{Plan-then-Write} mode for long-context tasks that demand structural integrity and a \textbf{Direct Generation} mode for short-context tasks that prioritize novelty (see examples in Figure~\ref{fig:long_short_case}). To train such a dual-mode system, we bypass the conventional Supervised Fine-Tuning (SFT) stage and introduce \textbf{Adaptive Constraint Preference Optimization (ACPO)}. This reference-free training paradigm leverages relative evaluation and self-bootstrapped baselines to optimize generation directly from user queries, eliminating the dependency on expensive human-written completions or ground-truth references.

Our contributions are threefold:
\begin{itemize}
    \item \textbf{Unified Framework:} We introduce UniCreative, the first framework to unify planning-based and intuition-driven creative generation within a single policy.
    \item \textbf{Reference-Free Optimization:} We propose ACPO, demonstrating that superior performance in both long-form and short-form regimes can be achieved solely through reinforcement learning, bypassing the need for SFT and ground-truth references.
    \item \textbf{Emergent Meta-cognition:} Extensive experiments demonstrate UniCreative's superior performance; crucially, our analysis reveals an emergent meta-cognitive ability to autonomously differentiate task regimes, validating the framework's scalability.
\end{itemize}

\begin{figure*}[h] 
  \includegraphics[width=\textwidth]{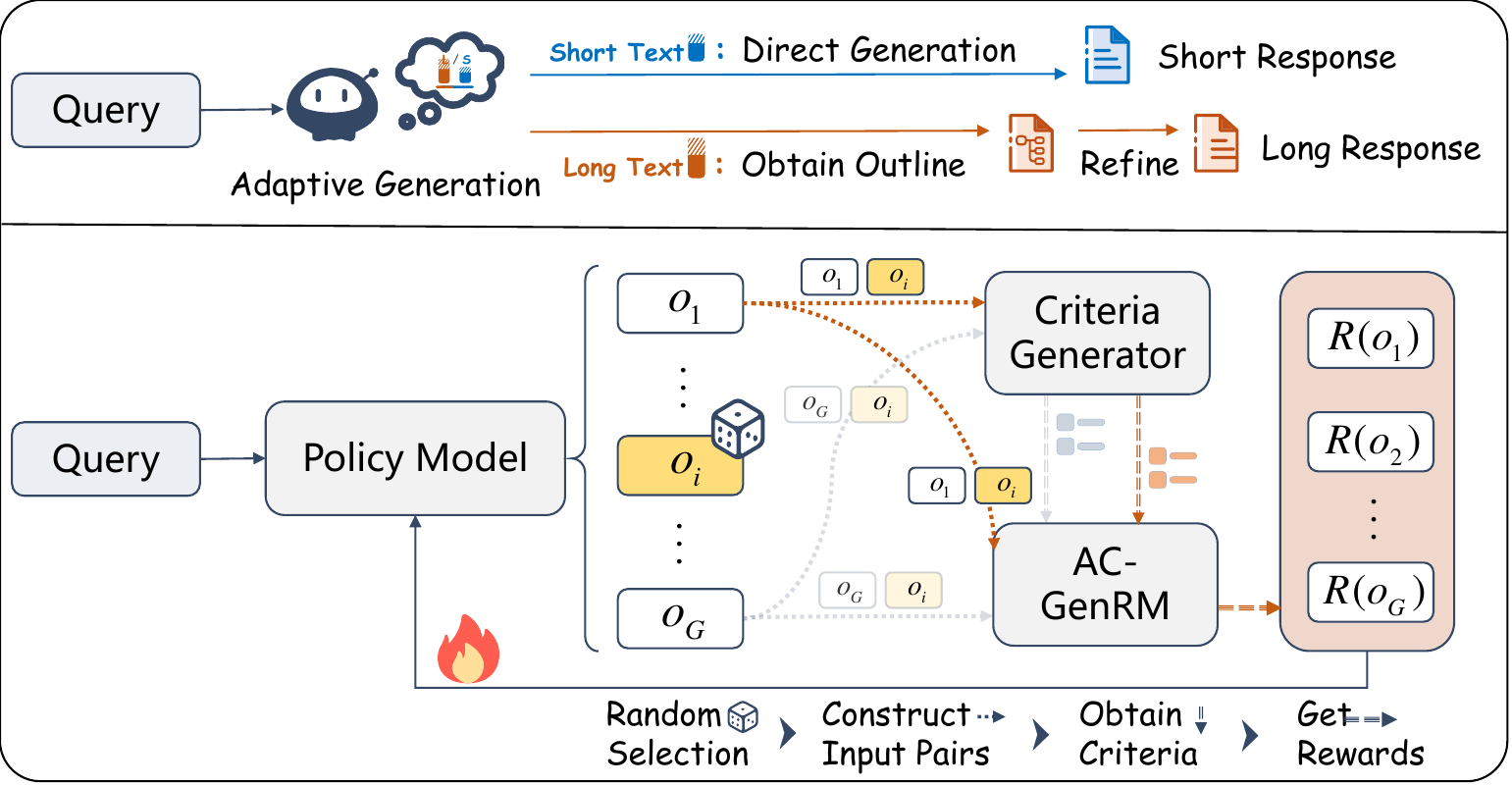} 
  \caption{\textbf{UniCreative Framework.} \textbf{Top:} The model adaptively selects \textit{Direct Generation} or \textit{Plan-then-Write} modes based on task nature. \textbf{Bottom:} ACPO training. The policy generates responses $\{o_1, \dots, o_G\}$ which are evaluated by \textbf{AC-GenRM} using dynamically synthesized criteria to provide reward signals for optimization.}
  \label{fig:flow_chart}
\end{figure*}

\section{Related Work}

\subsection{Creative Writing Evaluation and Benchmarks}
Evaluating creative writing is challenging due to its open-ended nature. While general benchmarks like LongBench focus on reasoning~\cite{bai2025longbench}, recent efforts target creativity specifically. WritingBench covers diverse genres~\cite{wu2025writingbench}, LitBench emphasizes expert-annotated preferences~\cite{fein2025litbench}, and CS4 measures creativity under constraints~\cite{atmakuru2024cs4}. Other benchmarks like UNCLE, LongWeave, and LongWriter-Zero further assess verifiability and RL challenges in ultra-long generation~\cite{yang2025uncle,xiao2025longweave,wu2025longwriter}.
To address the high cost of human evaluation, LLM-based evaluators (e.g., G-Eval) have been widely adopted~\cite{liu2023g}. However, these judges often exhibit biases or struggle with complex uncertainty assessments~\cite{atmakuru2024cs4,yang2025uncle}, motivating the need for more principled reward modeling.

\subsection{Structural Planning in Story Generation}
Explicit planning improves narrative coherence by separating high-level structure from surface realization~\cite{fan2018hierarchical,yao2019plan}. Despite the fluency of modern LLMs, they still suffer from ``myopia'' and inconsistency in long-form storytelling~\cite{xie2023next}. Consequently, structured guidance remains indispensable. Recent frameworks like ACE-RL and Beyond ReAct address this by integrating implicit planning or adaptive constraints, demonstrating significantly improved controllability in long-context generation~\cite{chen2025ace,wei2025beyond}.

\subsection{Preference Optimization and Generative Rewards}
Reinforcement learning from human feedback (RLHF) serves as the standard paradigm for aligning models with human preferences~\cite{christiano2017deep,ouyang2022training}, with Direct Preference Optimization (DPO) offering a streamlined, reward-free alternative~\cite{rafailov2023direct}. Recently, the focus has shifted toward enhancing the reward signal itself for open-ended generation. Generative reward models (GenRMs) reframe evaluation as a generative process, providing richer, chain-of-thought supervision compared to scalar scores~\cite{mahan2024generative,zhang2024generative}. Subsequent innovations have further augmented these models with reasoning capabilities (e.g., RM-R1, ReasonGRM) and improved generalization via inference-time scaling~\cite{chen2025rm,chen2025reasongrm,wang2025gram,liu2025inference}. In the specific domain of creative writing, methods such as Writing-Zero and others have begun to bridge the gap between non-verifiable creative objectives and RL, enabling effective optimization for long-form generation~\cite{lu2025writing,jia2025writing,liao2025rlmr,yu2025dapo}.

\section{Methodology}
\label{sec:methodology}

We reframe creative writing as a \textit{length-dependent adaptive decision process} requiring distinct strategies:
\begin{itemize}
    \item \textbf{Long-Form Narratives:} Require \textit{macroscopic structural integrity}. To counteract autoregressive ``myopia,'' the model employs a ``Plan-then-Write'' strategy, ensuring global coherence through hierarchical reasoning.
    \item \textbf{Short-Form Expressions:} Prioritize \textit{microscopic linguistic vibrancy}. Rigid planning here leads to ``over-determination,'' stifling the spontaneity and emotional resonance essential for creativity.
\end{itemize}

To dynamically decouple these pathways without human supervision, we propose a reference-free pipeline built on two pillars: (1) \textbf{AC-GenRM}, an evaluator providing instance-specific feedback; and (2) the \textbf{ACPO} algorithm, which optimizes the policy to autonomously select between planning and direct execution based on query complexity. The complete process is detailed in Figure~\ref{fig:flow_chart}.

\subsection{Adaptive Criteria GenRM (AC-GenRM)}
\label{sec:genrm}

Evaluating creative writing poses a significant challenge due to the subjective and multifaceted nature of literary quality. Standard reward models trained on general preference data often fail to capture the nuances of specific creative queries (e.g., distinguishing between ``suspense'' in a thriller and ``wit'' in a satire). Furthermore, LLM-based evaluators are notoriously prone to position bias. To address these limitations, we propose AC-GenRM, which decomposes the evaluation process into dynamic criteria synthesis and debiased pairwise ranking.

\paragraph{Dynamic Criteria Synthesis}
Instead of relying on a static system prompt for all queries, AC-GenRM first interprets the semantic intent of the input. Let $x$ be the user query. We model the evaluation criteria as a latent variable $C$. To equip the model with this capability, we employ Supervised Fine-Tuning (SFT) on a high-quality critique dataset, allowing the critic $\pi_{critic}$ to sample instance-specific criteria $C_x \sim \pi_{critic}(\cdot|x)$ that explicitly list the dimensions for assessment. For instance, given a prompt for a ``scary story,'' $C_x$ might prioritize ``plot twist'' and ``atmosphere,'' whereas for a ``greeting card,'' it would prioritize ``warmth'' and ``conciseness.'' Specific examples are shown in Figure~\ref{fig:criteria_case}. This dynamic generation ensures that the reward signal is aligned with the specific creative goals of the prompt.

\paragraph{Generative Reward Learning with Position Debiasing}
AC-GenRM is a generative judge trained via SFT to predict a winning label $l \in \{A, B\}$ given a query $x$, criteria $C_x$, and response pair $(y_A, y_B)$. To mitigate position bias, we employ Symmetrical Data Augmentation, swapping the response order with 50\% probability during training. The objective minimizes the negative log-likelihood of the correct label:
\begin{equation}
\begin{aligned}
\mathcal{L}_{RM}(\psi) = & -\mathbb{E}_{(x, C_x, y_A, y_B, l) \sim \mathcal{D}_{aug}} \\
& \log P_{\psi}(l \mid x, C_x, y_A, y_B)
\end{aligned}
\end{equation}
This forces the model to learn a position-invariant quality representation, ensuring preference signals strictly align with the synthesized criteria $C_x$.

\subsection{Adaptive Constraint Preference Optimization (ACPO)}
\label{sec:acpo}

With a robust evaluator in place, we introduce Adaptive Constraint Preference Optimization (ACPO). Unlike traditional RLHF which relies on a separate value model (critic), ACPO optimizes the policy $\pi_\theta$ directly using group-based relative feedback. This approach is particularly well-suited for long-context creative writing, where training a value model is computationally prohibitive and unstable.

\subsubsection{Reward Composition}
The reward function $R_{total}$ in ACPO is a composite signal designed to guide the model through the complex landscape of creative writing. It comprises three orthogonal components:

\paragraph{1. Bootstrapped Relative Reward ($R_{rel}$)}
We utilize a self-play mechanism to optimize without ground-truth references. For a query $x$, the policy generates $G$ responses $Y = \{y_1, \dots, y_G\}$. To resolve modality mismatch, we define a \textbf{projection operator} $\phi: \mathcal{V}^* \rightarrow \mathcal{V}^*$ that strips tokens in the planning segment $\mathcal{V}_{plan}$ (e.g., content between \texttt{<|plan|>} and \texttt{<|end\_plan|>}). For each $y_i$, we sample a baseline $y_{base} \in Y \setminus \{y_i\}$. 

The relative reward is derived from the discrete selection of the frozen generative AC-GenRM. Given $(x, C_x, \phi(y_i), \phi(y_{base}))$, the judge directly outputs the superior response, and the reward is assigned as:
\begin{equation}
R_{rel}(y_i) = 
\begin{cases} 
2, & \text{if AC-GenRM selects } \phi(y_i) \\ 
-2, & \text{otherwise} 
\end{cases}
\end{equation}
This binary signal creates a robust curriculum that continuously pushes the model's performance upper bound through internal competition.

\paragraph{2. Paradigm-Aware Structural Constraints ($R_{struct}$)}
To solve the ``Myopia'' problem in long texts and the ``Over-determination'' problem in short texts, we must enforce the correct cognitive mode. Let $\tau(x) \in \{\textsc{Long}, \textsc{Short}\}$ denote the task mode. We introduce an indicator function $h(y) = \mathbb{I}[\exists t \in y : t \in \mathcal{V}_{plan}]$ to detect planning actions.
To streamline the formulation, let $m_S = \mathbb{I}[\tau(x)=\textsc{Short}]$ and $m_L = \mathbb{I}[\tau(x)=\textsc{Long}]$ be the mode indicators. The structural penalty is defined as:
\begin{equation}
R_{struct}(y \mid x) = -\beta_{s} \cdot \big( m_S \cdot h(y) + m_L \cdot (1 - h(y)) \big)
\end{equation}
where $\beta_{s}$ is a penalty constant (set to 5.0 in our implementation). This term effectively prunes the exploration space, forcing the policy to collapse onto the structural paradigm appropriate for the task (i.e., mandating outlines for novels while prohibiting them for poems).

\paragraph{3. Adaptive Length Regularization ($R_{len}$)}
Structural adherence is a necessary but insufficient condition for high-quality generation. Empirically, we observe that RL policies often drift towards distinct failure modes: content collapse (insufficient length) in long-form narratives and verbosity (excessive length) in short-form expressions. To mitigate this, we introduce an asymmetric regularization mechanism with a penalty cap. Let $L(y) = |\phi(y)|$ denote the length of the generated content body. We formulate specific penalty functions:
\begin{align}
R_{long}(y) &= -\min \big( \lambda_{l} \cdot \text{ReLU}(\theta_{min} - L(y)), \gamma \big) \\
R_{short}(y) &= -\min \big( \lambda_{s} \cdot \text{ReLU}(L(y) - \theta_{max}), \gamma \big)
\end{align}
where $\gamma$ is the maximum penalty cap, set to $5.0$ in our implementation to prevent excessive gradient signals from outlier lengths. Consequently, the final regularization term $R_{len}(y \mid x)$ dynamically activates either $R_{long}$ or $R_{short}$ contingent upon $\tau(x)$, creating a soft boundary for appropriate information density.

\subsubsection{Optimization Objective}

We aggregate the rewards into a total score $R_{total}(y_i) = R_{rel} + R_{struct} + R_{len}$. To stabilize training without a value model, we employ \textbf{Group Relative Policy Optimization (GRPO)}. GRPO computes the advantage $A_i$ by normalizing the rewards within the sampled group $Y$, using the group mean as the baseline:
\begin{equation}
A_i = \frac{R_{total}(y_i) - \mu(\{R_{total}(y_j)\}_{j=1}^G)}{\sigma(\{R_{total}(y_j)\}_{j=1}^G) + \epsilon}
\end{equation}
This group-based normalization significantly reduces the variance of the gradient estimate. 
To streamline the objective formulation, we define the clipped advantage term $\mathcal{L}^{clip}_i$ and the KL divergence term $\mathcal{D}^{KL}_i$ as follows:
\begin{align}
\mathcal{L}^{clip}_i &= \min \big( \rho_i A_i, \text{clip}(\rho_i, 1-\epsilon, 1+\epsilon) A_i \big) \nonumber \\
\mathcal{D}^{KL}_i &= D_{KL}(\pi_\theta(y_i|x) \parallel \pi_{ref}(y_i|x)) \nonumber
\end{align}
Consequently, the final objective function maximizes:
\begin{equation}
\mathcal{J}(\theta) = \mathbb{E}_{x, Y} \Bigg[ \frac{1}{G} \sum_{i=1}^G \Big( \mathcal{L}^{clip}_i - \beta_{KL} \cdot \mathcal{D}^{KL}_i \Big) \Bigg]
\end{equation}
where $\mathbb{E}_{x, Y}$ denotes the expectation over the dataset and policy rollouts, $\rho_i = \frac{\pi_\theta(y_i|x)}{\pi_{\theta_{old}}(y_i|x)}$ is the importance sampling ratio, and $\beta_{KL}$ prevents excessive deviation from the reference model $\pi_{ref}$, ensuring linguistic fluency.

\section{Experiments}

\subsection{Task Design and Data Construction}

To train the UniCreative framework, we formalize creative writing into two distinct generative paradigms and construct a composite training corpus derived from diverse sources.

\paragraph{1. Definition of Generative Paradigms}
We decouple the conditional generation problem $P(y|x)$ into two specialized sub-tasks:
\begin{itemize}
    \item \textbf{Planning-Augmented Generation (for Long-Form):} Designed for tasks requiring structural depth (e.g., novels). The model is tasked to first generate an intermediate latent variable $z$ (representing the outline or reasoning chain) before producing the final text $y$. The sequence is formalized as: $x \rightarrow \texttt{<|plan|>} \ z \ \texttt{<|end\_plan|>} \rightarrow y$. This explicitly compels the model to engage in macroscopic reasoning.
    \item \textbf{Direct Generation (for Short-Form):} Designed for tasks requiring high entropy (e.g., poetry, slogans). The model must map $x \rightarrow y$ directly, with a strict prohibition on the \texttt{<|plan|>} token. This preserves the stochasticity and spontaneity of the token distribution.
\end{itemize}

\paragraph{2. Training Data Construction}
We construct a composite bilingual dataset for reward modeling and reference-free reinforcement learning:

\begin{itemize}
    \item \textbf{Preference Data for AC-GenRM:} We utilize LitBench~\cite{fein2025litbench} (43k samples) to capture long-form narrative preferences and the Blessing dataset~\cite{wei2025igniting} (6k samples) for short-form emotional expressions.
    
    \item \textbf{Exploration Prompts for ACPO:} We aggregate a diverse query set for the exploration space. The Long-Context Corpus combines HelloBench~\cite{que2024hellobench}, the Story Writing Benchmark, and high-quality web fiction, further augmented via the WritingBench~\cite{wu2025writingbench} synthesis methodology. The Short-Context Corpus consists of curated online queries designed to train expressive tension within constrained lengths.
\end{itemize}

Crucially, ACPO training relies solely on these queries without ground-truth completions, allowing the model to learn autonomous paradigm switching driven purely by AC-GenRM feedback.

\subsection{Training Configuration}
We trained the Qwen3 series models (1.7B, 4B, and 8B) on 8 NVIDIA H800 GPUs. Detailed training configurations are provided in the Appendix~\ref{sec:training_config}.


\definecolor{graybg}{gray}{0.95}

\begin{table}[h]
\centering
\small
\renewcommand{\arraystretch}{1.15} 
\setlength{\tabcolsep}{3.5pt} 

\resizebox{\linewidth}{!}{
\begin{tabular}{llcc}
\toprule
\textbf{Category} & \textbf{Model} & \textbf{LitBench} & \textbf{Blessing} \\
\midrule
\multirow{6}{*}{\textit{Baselines}} 
& Claude-Sonnet-3.7 & 0.731 & 0.902 \\
& Claude-3.5-Haiku  & 0.675 & 0.949 \\
& GPT-4.1           & 0.702 & 0.923 \\
& GPT-4.1-Mini      & 0.630 & 0.969 \\
& DeepSeek-V3       & 0.700 & 0.906 \\
& DeepSeek-R1       & 0.710 & 0.977 \\
\midrule

\rowcolor{graybg}
 & Base & 0.477 & 0.643 \\
\rowcolor{graybg}
\multirow{-3}{*}{\textbf{Qwen3-0.6B}} & AC-GenRM & 0.728 & 0.989 \\

 & Base & 0.543 & 0.849 \\
\multirow{-3}{*}{\textbf{Qwen3-1.7B}} & AC-GenRM & 0.776 & 0.991 \\

\rowcolor{graybg}
 & Base & 0.666 & 0.938 \\
\rowcolor{graybg}
\multirow{-3}{*}{\textbf{Qwen3-4B}} & AC-GenRM & 0.796 & 0.993 \\

 & Base & 0.688 & 0.944 \\
\multirow{-3}{*}{\textbf{Qwen3-8B}} & \textbf{AC-GenRM} & \textbf{0.807} & \textbf{0.994} \\

\bottomrule
\end{tabular}
}
\caption{Reward model evaluation showing agreement rates with proprietary judges on \textbf{LitBench} and \textbf{Blessing}. AC-GenRM is compared against unaligned Base models to isolate the algorithmic contribution.}
\label{tab:genrm_results}
\end{table}

\begin{table*}[h!]
\centering

\resizebox{\textwidth}{!}{%
\begin{tabular}{lccccccccccccc}
\toprule
\textbf{Models} & \textbf{Avg} & \textbf{D1} & \textbf{D2} & \textbf{D3} & \textbf{D4} & \textbf{D5} & \textbf{D6} & \textbf{R1} & \textbf{C} & \textbf{R2} & \textbf{C} & \textbf{R3} & \textbf{C} \\
\midrule
\multicolumn{14}{l}{\textit{\textbf{Proprietary LLMs}}} \\
\midrule
O3-2025-04-16 & 85.27 & 84.81 & 85.20 & 83.89 & 85.88 & 85.82 & 86.80 & 85.14 & 87.45 & 85.24 & 90.98 & 86.31 & 87.21 \\
Gemini-2.5-pro-preview & 83.05 & 83.21 & 81.47 & 83.00 & 84.52 & 84.49 & 82.14 & 83.58 & 86.49 & 83.86 & 90.45 & 83.35 & 83.98 \\
Claude-Sonnet-3.7 & 78.48 & 78.24 & 77.93 & 76.51 & 79.37 & 79.26 & 80.88 & 79.43 & 82.51 & 78.84 & 86.14 & 79.23 & 80.49 \\
GPT-4o & 75.46 & 74.40 & 73.42 & 74.38 & 77.91 & 75.86 & 78.08 & 76.82 & 81.57 & 75.82 & 85.46 & 76.13 & 76.73 \\
o1-Preview & 68.57 & 68.54 & 67.01 & 66.57 & 69.53 & 70.31 & 71.41 & 70.09 & 75.10 & 68.49 & 79.78 & 70.91 & 73.81 \\
\midrule
\multicolumn{14}{l}{\textit{\textbf{Open-source LLMs}}} \\
\midrule
DeepSeek-R1-0528 & 83.22 & 83.15 & 81.48 & 81.55 & 85.68 & 84.14 & 84.44 & 84.24 & 87.27 & 83.72 & 89.35 & 83.83 & 82.70 \\
Qwen3-235B-A22B-Thinking & 81.45 & 80.19 & 79.24 & 80.95 & 82.92 & 82.52 & 82.89 & 82.54 & 85.02 & 81.30 & 88.22 & 81.26 & 81.76 \\
LongWriter-Zero-32B & 80.30 & 80.66 & 80.27 & 80.21 & 76.09 & 83.55 & 81.02 & 79.89 & 83.38 & 80.77 & 86.82 & 80.23 & 82.05 \\
Qwen3-8B-Thinking & 75.56 & 75.47 & 74.42 & 75.51 & 73.03 & 77.80 & 77.16 & 76.29 & 79.72 & 75.67 & 84.96 & 74.35 & 77.30 \\
Qwen3-235B-A22B & 73.63 & 73.56 & 72.90 & 73.98 & 70.13 & 76.52 & 74.69 & 77.46 & 82.05 & 77.01 & 87.29 & 76.26 & 79.57 \\
Qwen3-8B & 70.75 & 70.68 & 70.74 & 70.33 & 68.51 & 72.74 & 71.52 & 71.72 & 76.77 & 71.69 & 83.12 & 70.06 & 75.36 \\
Qwen-2.5-72B-instruct & 65.28 & 65.80 & 63.36 & 63.80 & 62.75 & 68.07 & 67.91 & 65.81 & 70.49 & 65.92 & 78.65 & 66.38 & 67.95 \\
LongWriter-glm-9B & 62.94 & 64.06 & 63.66 & 62.35 & 61.26 & 65.03 & 61.30 & 62.79 & 66.70 & 63.60 & 74.82 & 63.37 & 65.88 \\
LongWriter-llama3.1-8B & 58.01 & 60.05 & 59.31 & 57.58 & 56.03 & 58.38 & 56.73 & 58.12 & 61.40 & 58.60 & 67.61 & 59.05 & 62.97 \\
Llama-3.3-70B-instruct & 50.43 & 50.67 & 49.25 & 47.90 & 48.52 & 52.92 & 56.56 & 50.71 & 50.71 & 50.38 & 50.38 & 51.08 & 51.08 \\
\midrule
\multicolumn{14}{l}{\textit{\textbf{Our Models}}} \\
\midrule
Qwen3-1.7B-Thinking & 63.06 & 66.72 & 64.41 & 64.13 & 58.13 & 60.52 & 63.67 & 64.92 & 70.02 & 62.96 & 70.43 & 61.16 & 60.53 \\
Qwen3-1.7B-Thinking + RL & 73.65 & 76.06 & 74.54 & 74.03 & 68.54 & 75.14 & 74.50 & 74.71 & 78.80 & 74.43 & 81.59 & 73.31 & 73.67 \\
Qwen3-4B-Thinking & 71.35 & 72.57 & 73.10 & 72.06 & 64.38 & 74.54 & 72.95 & 72.08 & 76.79 & 72.17 & 79.18 & 70.51 & 70.93 \\
Qwen3-4B-Thinking + RL & 77.36 & 78.63 & 78.48 & 77.88 & 72.61 & 79.55 & 77.91 & 77.83 & 82.01 & 78.20 & 85.62 & 76.71 & 77.92 \\
Qwen3-8B-Thinking & 77.11 & 77.54 & 77.73 & 77.63 & 73.68 & 78.49 & 78.42 & 77.62 & 81.79 & 77.74 & 83.60 & 76.58 & 76.86 \\
Qwen3-8B-Thinking + RL & \textbf{82.42} & \textbf{83.17} & \textbf{81.85} & \textbf{83.05} & \textbf{80.80} & \textbf{83.57} & \textbf{82.72} & \textbf{82.68} & \textbf{85.32} & \textbf{82.68} & \textbf{87.61} & \textbf{81.60} & \textbf{80.62} \\
\bottomrule
\end{tabular}%
}
\caption{Performance of different LLMs on WritingBench across six domains and three writing requirements. Scores are normalized from a 0-10 range to a 100-point scale. The corresponding domains and requirements are: (D1) Academic \& Engineering, (D2) Finance \& Business, (D3) Politics \& Law, (D4) Literature \& Art, (D5) Education, (D6) Advertising \& Marketing, (R1) Style, (R2) Format, and (R3) Length. ``C'' denotes the category-specific scores of the three requirements.}
\label{tab:long_results}
\end{table*}

\subsection{Evaluation Benchmarks}
\label{sec:benchmarks}

We evaluate UniCreative using a multi-tiered benchmark suite covering reward modeling, generative quality, and meta-cognitive decision-making.
Specifically, we assess (i) reward model quality via agreement with strong LLM judges on preference-based creative writing benchmarks, (ii) long-form and short-form generative performance using established creative writing datasets, and (iii) the model’s ability to autonomously distinguish between planning-intensive and direct generation tasks.
Detailed benchmark descriptions are provided in Appendix~\ref{sec:benchmark_details}.



\begin{table*}[t]
\centering
\small
\renewcommand{\arraystretch}{1.2} 
\setlength{\tabcolsep}{8pt} 

\begin{tabular}{lc |lcc} 
\toprule
\multicolumn{2}{c}{\textbf{Reference LLMs}} & \multicolumn{3}{c}{\textbf{Our Models (Ablation Study)}} \\
\cmidrule(r){1-2} \cmidrule(l){3-5}
\textbf{Model Name} & \textbf{Score} & \textbf{Model Name} & \textbf{Score} & \textbf{Gain ($\Delta$)} \\
\midrule
DeepSeek-V3-250324 & 87.2\% & Qwen3-1.7B & 8.40\% & - \\
DeepSeek-V3.2 & 90.4\% & Qwen3-1.7B-Thinking & 64.2\% & - \\
Qwen3-32B & 45.2\% & \textbf{Qwen3-1.7B-Thinking + RL} & \textbf{90.0\%} & \textcolor{teal}{+25.8\%} \\
Qwen3-32B-Thinking & 88.4\% & Qwen3-4B & 40.8\% & - \\
GPT-4.1 & 89.4\% & Qwen3-4B-Thinking & 74.0\% & - \\
Doubao-Seed-1.6 & \textbf{94.6\%} & \textbf{Qwen3-4B-Thinking + RL} & \textbf{91.4\%} & \textcolor{teal}{+17.4\%} \\
Grok-4 & 80.6\% & Qwen3-8B & 43.6\% & - \\
Claude-Sonnet-4 & 92.8\% & Qwen3-8B-Thinking & 68.0\% & - \\
Claude-Sonnet-4.5 & 93.2\% & \textbf{Qwen3-8B-Thinking + RL} & \textbf{93.6\%} & \textbf{\textcolor{teal}{+25.6\%}} \\

\bottomrule
\end{tabular}
\caption{Performance evaluation on the Blessing short-text dataset. We employ an Adversarial Framework to benchmark standard LLMs (left) against our UniCreative variants (right). \textbf{Score} represents the percentage of excellent ratings. The \textbf{Gain ($\Delta$)} column highlights the significant score improvement achieved by our RL method compared to the `thinking` baseline.}
\label{tab:short_results}
\end{table*}


\begin{table*}[t]
\centering
\small 
\setlength{\tabcolsep}{10pt}
\renewcommand{\arraystretch}{1.4}

\begin{tabular}{llcccc}
\toprule
\multirow{2}{*}{\textbf{Model Size}} & \multirow{2}{*}{\textbf{Method}} & \multicolumn{4}{c}{\textbf{Accuracy on Mode Discrimination (5-run Avg. $\pm$ SD)}} \\
\cmidrule(l){3-6}
 & & \textbf{Easy} & \textbf{Hard} & \textbf{Avg.} & \textbf{Gain ($\Delta$)} \\
\midrule

\multirow{2}{*}{Qwen3-1.7B} 
 & Thinking & 64.2\% {\scriptsize $\pm$ 3.1\%} & 65.1\% {\scriptsize $\pm$ 2.2\%} & 64.7\% & - \\
 & + RL (Ours) & 57.0\% {\scriptsize $\pm$ 1.1\%} & 56.6\% {\scriptsize $\pm$ 1.4\%} & 56.8\% & \textcolor{teal}{-7.9\%} \\
\cmidrule(l){2-6} 

\multirow{2}{*}{Qwen3-4B} 
 & Thinking & 77.1\% {\scriptsize $\pm$ 2.2\%} & 62.4\% {\scriptsize $\pm$ 2.1\%} & 69.8\% & - \\
 & + RL (Ours) & 95.9\% {\scriptsize $\pm$ 0.7\%} & 82.8\% {\scriptsize $\pm$ 1.7\%} & 89.4\% & \textbf{\textcolor{teal}{+19.6\%}} \\
\cmidrule(l){2-6}

\multirow{2}{*}{Qwen3-8B} 
 & Thinking & 80.5\% {\scriptsize $\pm$ 2.1\%} & 68.5\% {\scriptsize $\pm$ 1.9\%} & 74.5\% & - \\
 & + RL (Ours) & \textbf{96.4\%} {\scriptsize $\pm$ 1.8\%} & \textbf{86.4\%} {\scriptsize $\pm$ 1.2\%} & \textbf{91.4\%} & \textbf{\textcolor{teal}{+16.9\%}} \\

\bottomrule
\end{tabular}
\caption{Accuracy on the Mode Discrimination Benchmark across different model scales (5-run avg). Significant gains in 4B and 8B models underscore the efficacy of ACPO in fostering emergent meta-cognitive planning.}
\label{tab:scaling_law}
\end{table*}

\section{Results and Analysis}

\subsection{Justification for Pairwise Generative Reward Modeling}
\label{sec:why_pairwise_genrm}

Our AC-GenRM architecture departs from traditional pointwise scalar models and discriminative Bradley-Terry (BT) classifiers. Motivated by recent findings in creative evaluation~\citep{wu2025writingbench,fein2025litbench,jia2025writing}, we adopt a pairwise generative approach for two primary reasons.

\noindent \textbf{Pairwise Comparison over Pointwise Scoring.} Unlike math or code, creative writing lacks definitive ground truth. Pointwise scoring suffers from calibration instability, where absolute scores (e.g., ``8/10'') remain ambiguous without a fixed reference. In contrast, relative preference ($y_A \succ y_B$) aligns better with human intuition and reduces variance driven by prompt difficulty~\citep{christiano2017deep}. Moreover, comparative discrimination mitigates the reward hacking common in scalar models—such as verbosity bias—thereby improving generalization across heterogeneous tasks.

\noindent \textbf{Generative Reasoning vs. Discriminative BT.} Unlike opaque discriminative models, AC-GenRM leverages generative reasoning to synthesize query-specific criteria ($C_x$), ensuring precise semantic alignment with user intent across diverse tasks. This self-principled critique mechanism enhances robustness against reward hacking, with SFT on preference data serving as a crucial stabilization mechanism for the reasoning process. Empirically, AC-GenRM yields superior signal quality, achieving an 80.7\% agreement rate that significantly surpasses Claude-3.7-Sonnet (73.1\%) and exceeds the performance of top-tier discriminative models ($\sim$78\%) on the same benchmark.

\subsection{Performance on Long-Form and Short-Form Writing}

We evaluate UniCreative on complex narratives (\textit{WritingBench}) and high-entropy short-form text (\textit{Blessing}).

\noindent \textbf{Long-Form Performance (WritingBench).} 
As shown in Table~\ref{tab:long_results}, ACPO consistently enhances long-context reasoning across all scales. \texttt{Qwen3-8B-Thinking + RL} achieved an average score of \textbf{82.42} (+5.31 over the Base model), significantly outperforming much larger models like \texttt{Llama-3.3-70B-Instruct} (50.43) and \texttt{Qwen-2.5-72B-Instruct} (65.28), while rivaling \texttt{Claude-Sonnet-3.7} (78.48). 
Specifically, ACPO improved compliance in Format (R2) and Length (R3). Unlike specialized baselines like ACE-RL \citep{chen2025ace} that rely on rigid, task-specific checklists, UniCreative maintains structural integrity without sacrificing the flexibility needed for broader creative tasks.

\noindent \textbf{Short-Form Performance (Blessing).} 
Table~\ref{tab:short_results} demonstrates the critical necessity of mode switching. The ``Thinking'' baselines struggle with short-form tasks (e.g., 64.2\% for 1.7B) due to ``over-determination'' from forced planning. 
By enabling the model to bypass planning, \textbf{ACPO achieves massive gains}: \texttt{Qwen3-1.7B + RL} jumped to 90.0\% (+25.8\%), matching \texttt{DeepSeek-V3.2}, while the 8B variant reached \textbf{93.6\%}, surpassing \texttt{Claude-Sonnet-4.5} (93.2\%). These results confirm that our reference-free RL enables the model to ``unlearn'' rigid structural constraints when appropriate, restoring the linguistic vibrancy essential for high-quality short-form generation.

\subsection{Can the Model Adaptively Differentiate Task Regimes?}

We evaluate whether UniCreative enables models to distinguish between generation pathways using the Mode Discrimination Benchmark.

\noindent \textbf{Scalable Accuracy Gains.}
As shown in Table~\ref{tab:scaling_law}, ACPO significantly improves mode-switching accuracy in larger models. \texttt{Qwen3-4B} and \texttt{Qwen3-8B} achieve up to 96.4\% and 86.4\% accuracy on Easy and Hard tasks, respectively. This confirms that given sufficient capacity, our framework effectively aligns the policy with structural preferences, allowing models to autonomously invoke the optimal generation mode.

\noindent \textbf{Capacity Bottleneck in Small Models.}
Conversely, \texttt{Qwen3-1.7B} exhibits a decline in discrimination accuracy (e.g., 64.2\% $\rightarrow$ 57.0\%). This does not signify training failure; rather, it reflects a \textit{capacity bottleneck}. While RL improves the 1.7B model's raw writing quality (Table~\ref{tab:long_results} and Table~\ref{tab:short_results}), its limited parameter count forces a trade-off: the model prioritizes high-entropy linguistic vibrancy over the complex meta-cognitive logic required for autonomous paradigm switching.

\noindent \textbf{Emergent Meta-Cognition.}
These results underscore an emergent scaling trend. While the 1.7B model yields to the tension between content and structure, the 4B and 8B models successfully resolve it. The ability to identify latent structural complexity and invoke appropriate computational pathways constitutes a high-order \textit{meta-cognitive reasoning} that only stabilizes once a specific model scale threshold is exceeded.


\section{Conclusion}

We introduced \textbf{UniCreative}, a unified RL framework balancing long-form logical coherence with short-form expressive vibrancy. By leveraging \textbf{AC-GenRM} for criteria synthesis and \textbf{ACPO} for reference-free optimization, our approach enables models to autonomously navigate planning versus direct generation trade-offs without SFT or expensive annotations. Results show UniCreative mitigates generative failures and fosters an emergent meta-cognitive ability to select optimal computational pathways based on task complexity. This framework establishes a robust foundation for aligning LLMs with the nuanced demands of creative writing.

\section*{Limitations}

Despite the strong performance and emergent meta-cognitive abilities demonstrated by \textsc{UniCreative}, several limitations remain that offer avenues for future research:

\paragraph{Dependency on Model Scale} 
Our analysis of the Mode Discrimination Benchmark (Table~\ref{tab:scaling_law}) reveals a clear capacity bottleneck in smaller models. While meta-cognitive task differentiation emerges at the 4B and 8B scales, the 1.7B model struggles to balance structural logic with linguistic vibrancy. This suggests that the proposed adaptive switching mechanism may require a minimum parameter threshold to stabilize, potentially limiting its effectiveness for ultra-lightweight edge deployment.

\paragraph{Boundary Cases in Task Regimes} 
UniCreative currently operates on a binary switch between \textsc{Long-form} (Plan-then-Write) and \textsc{Short-form} (Direct Generation). However, there exists a "medium-form" gray area—such as long social media threads or short analytical essays—where a rigid plan might be too heavy, yet direct generation might lack sufficient coherence. Future work could explore a more granular or "soft" planning mechanism that adaptively adjusts the density of the reasoning chain based on the prompt's intermediate requirements.

\paragraph{Computational Overhead of Long-Context RL} 
Optimizing policies for long-form narratives (up to 12.6k tokens per sample) is computationally intensive. Although we utilized Group Relative Policy Optimization (GRPO) to bypass the value model and save VRAM, the exploration of such a vast policy space requires significant GPU resources (e.g., 8 $\times$ NVIDIA H800) and training time. This overhead may pose challenges for academic labs with constrained hardware budgets looking to replicate or extend ultra-long-context training.


\bibliography{main}

\appendix

\section{Appendix}
\label{sec:appendix}

\subsection{Training Configuration}
\label{sec:training_config}

Our experiments are conducted on a computational cluster equipped with 8 $\times$ NVIDIA H800 (80GB) GPUs. The training pipeline consists of two sequential stages: supervised fine-tuning for the reward model and reinforcement learning for policy optimization.

\paragraph{GenRM Supervised Fine-Tuning}
We initialize the reward model using the Qwen3-8B backbone. To transform the generative model into a robust judge, we perform full-parameter fine-tuning on the constructed GenRM dataset. Given the memory demands of full-parameter updates, we leverage DeepSpeed ZeRO-3 offloading strategies. The model is trained for 2 epochs with a conservative learning rate of $1\times 10^{-5}$ and a global batch size of 256. We utilize a cosine learning rate scheduler with a 10\% warmup ratio. The maximum context length is set to 5,200 tokens to accommodate the query, response, and the generated evaluation rationale.

\paragraph{ACPO Reinforcement Learning}
To address the computational bottleneck of processing extremely long contexts (up to 12.6k tokens per sample), we integrate vLLM for efficient rollout generation and employ Tensor Parallelism (TP=2) to distribute the model across GPUs during inference. The training utilizes a global batch size of 1024, spanning 10 epochs with a conservative learning rate of $1\times 10^{-6}$. We enforce strict length constraints with a maximum prompt length of 8,600 tokens and a response limit of 4,000 tokens to accommodate extensive planning chains. For each query, we sample $G=6$ distinct responses to compute the group-based advantage. To ensure training stability under memory constraints, we enable gradient checkpointing and offload optimizer states to the CPU via FSDP. Furthermore, we apply a low-variance KL penalty ($\beta_{KL}=0.001$) and disable entropy regularization to prevent policy collapse while maintaining alignment with the reference model.

\begin{figure*}[h!]
    \centering
    
    \begin{subfigure}[b]{0.32\textwidth}
        \includegraphics[width=\linewidth]{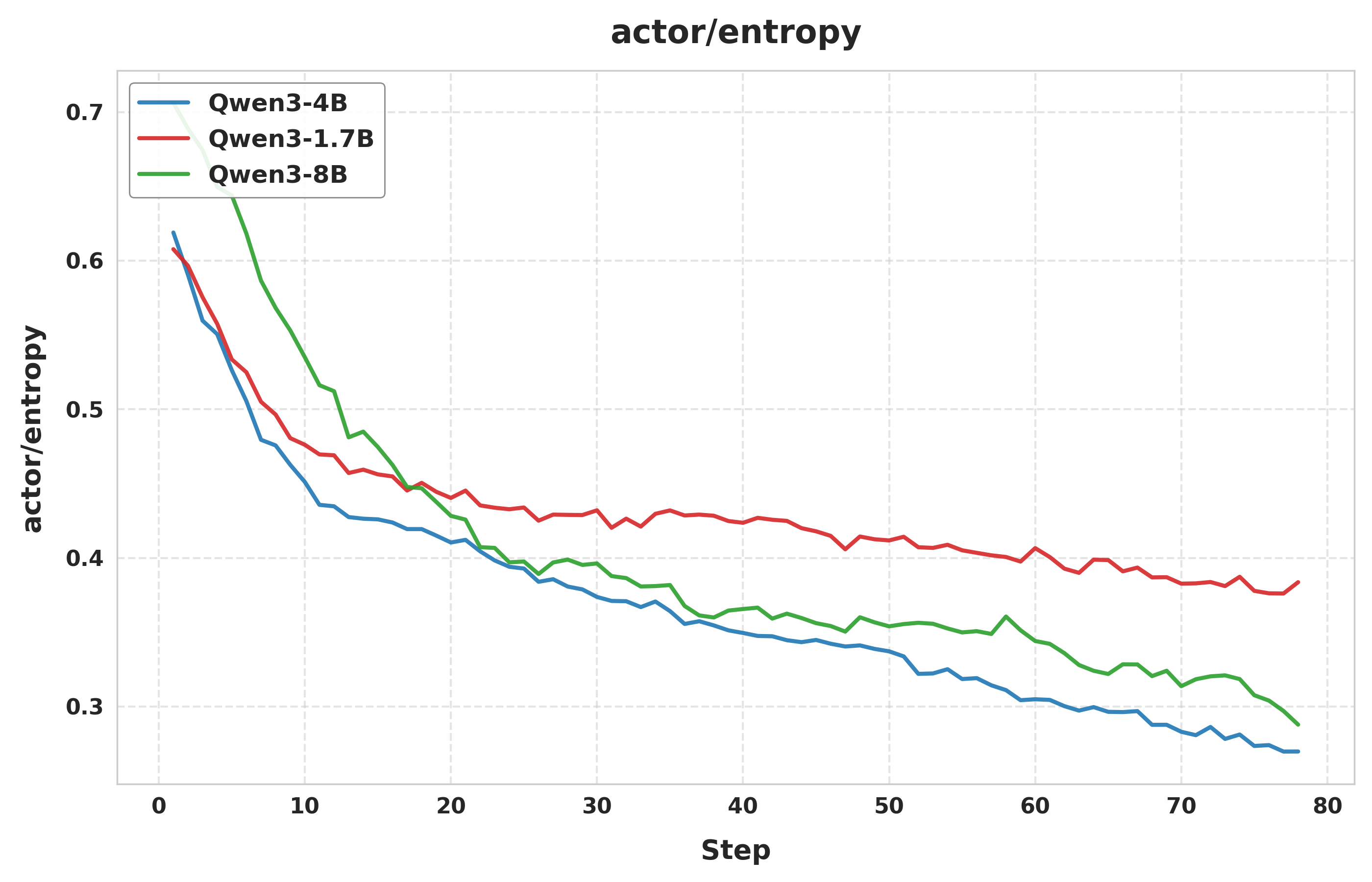}
        \caption{Metric 1}
    \end{subfigure}
    \hfill
    \begin{subfigure}[b]{0.32\textwidth}
        \includegraphics[width=\linewidth]{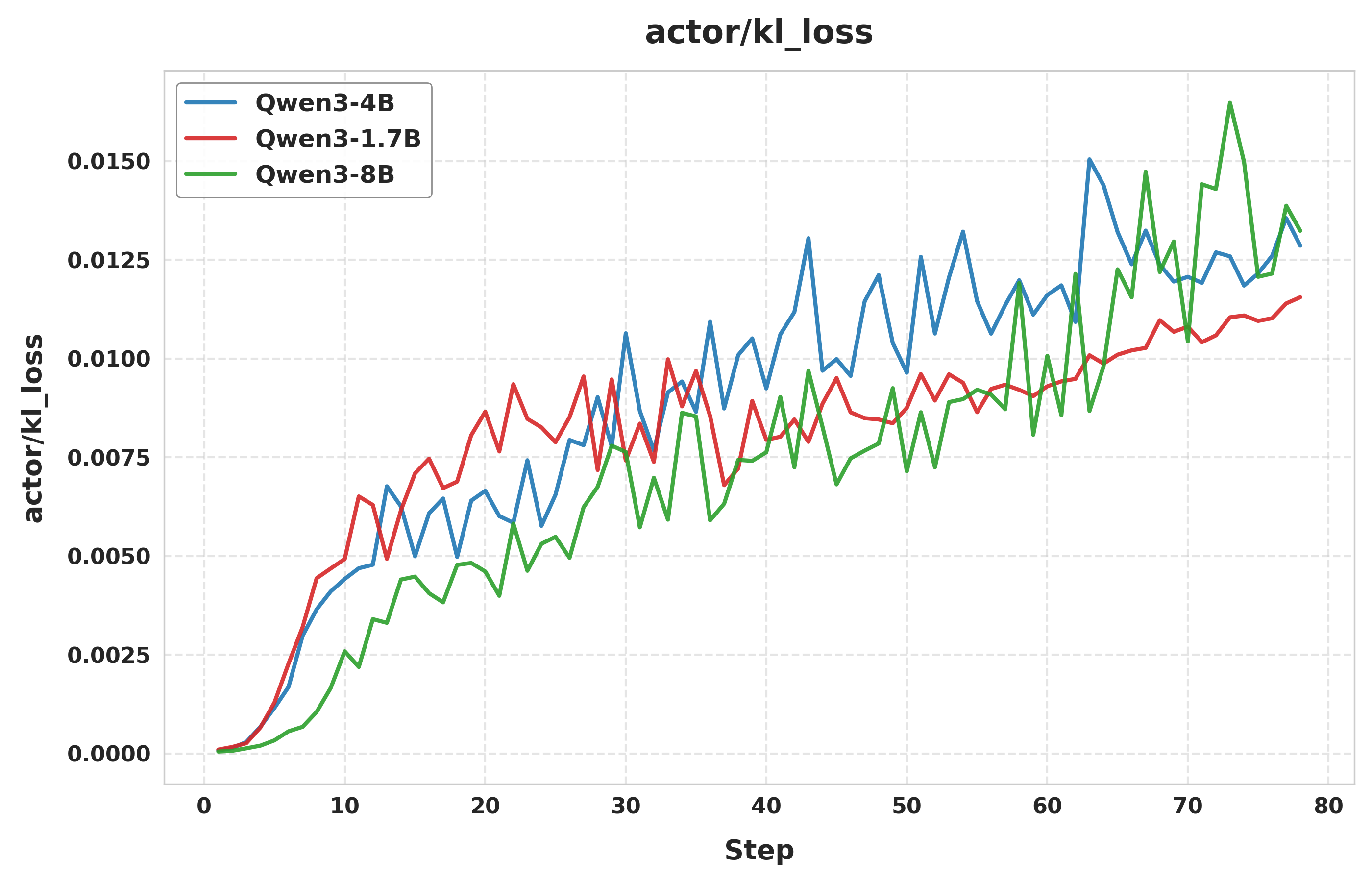}
        \caption{Metric 2}
    \end{subfigure}
    \hfill
    \begin{subfigure}[b]{0.32\textwidth}
        \includegraphics[width=\linewidth]{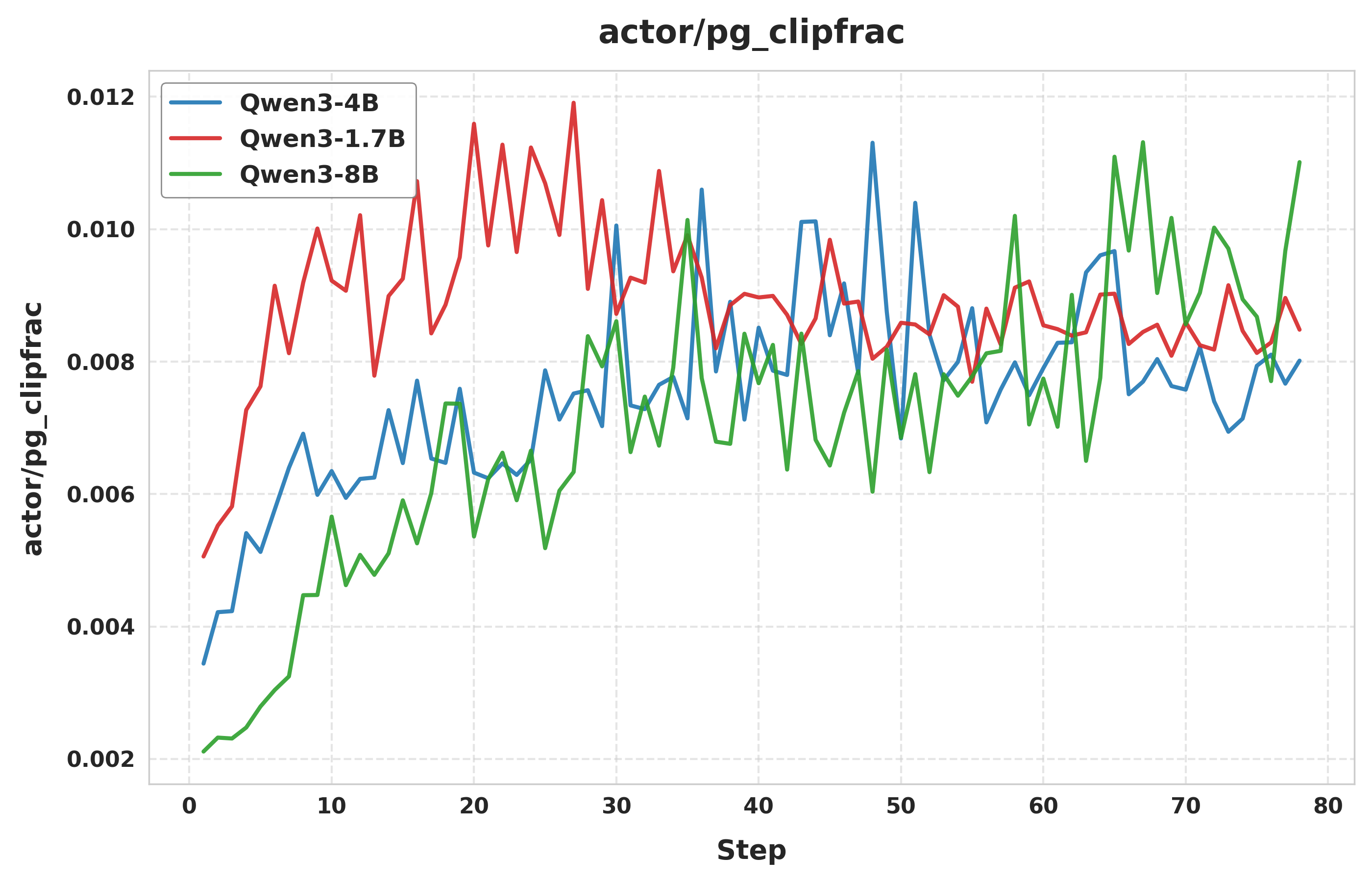}
        \caption{Metric 3}
    \end{subfigure}
    
    \vspace{0.5cm}

    \begin{subfigure}[b]{0.32\textwidth}
        \includegraphics[width=\linewidth]{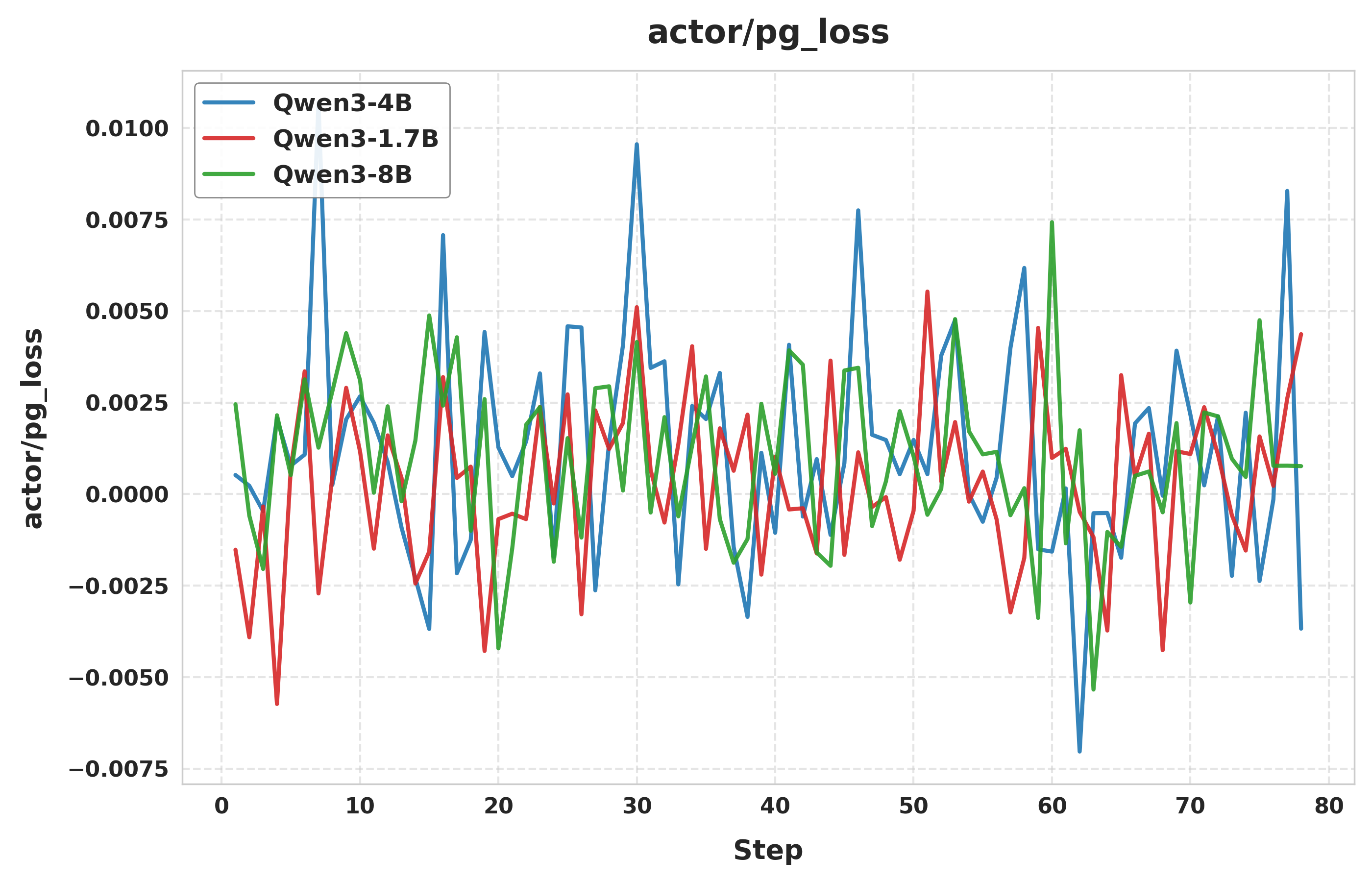}
        \caption{Metric 4}
    \end{subfigure}
    \hfill
    \begin{subfigure}[b]{0.32\textwidth}
        \includegraphics[width=\linewidth]{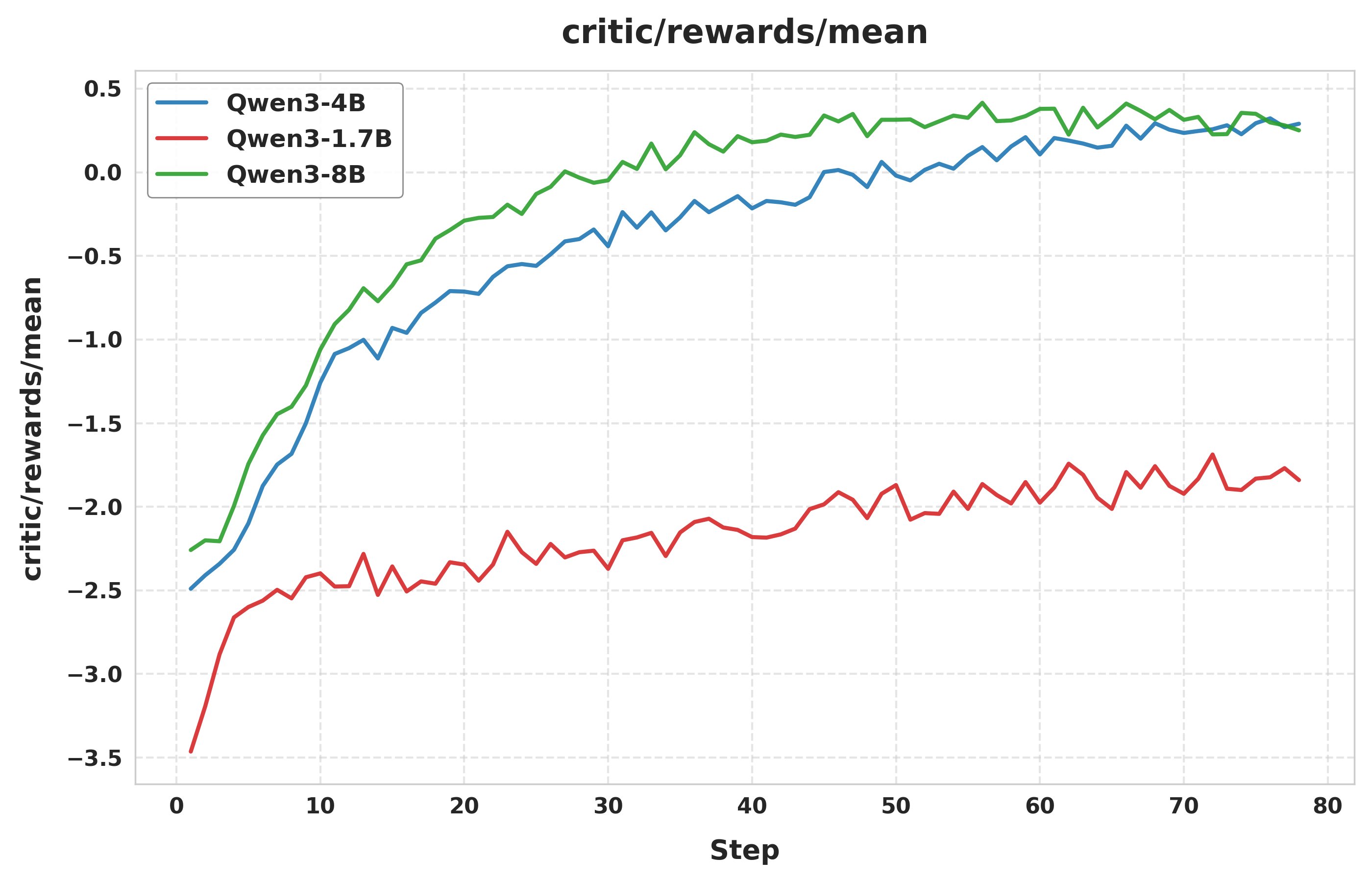}
        \caption{Metric 5}
    \end{subfigure}
    \hfill
    \begin{subfigure}[b]{0.32\textwidth}
        \includegraphics[width=\linewidth]{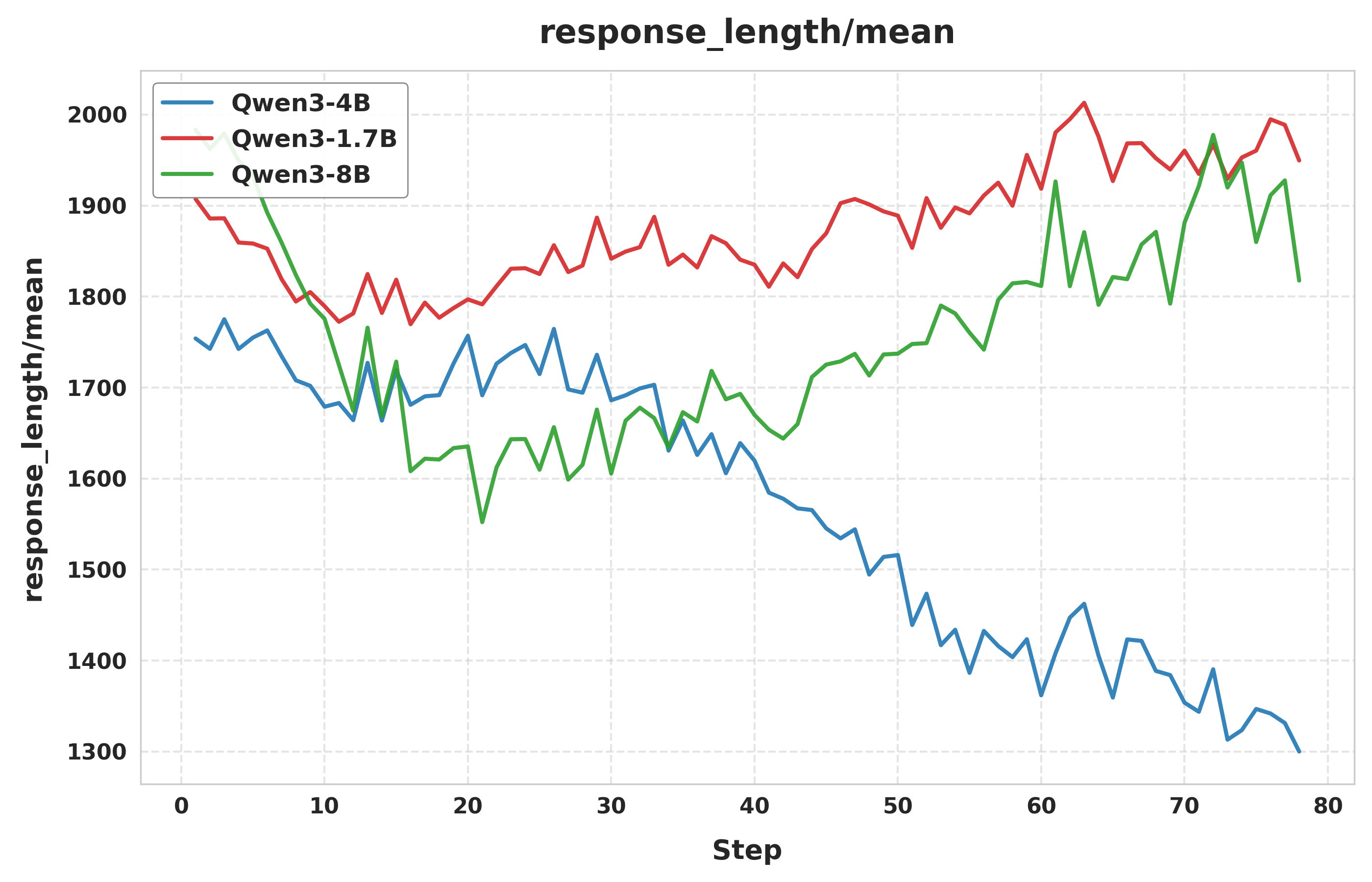}
        \caption{Metric 6}
    \end{subfigure}
    \caption{\textbf{Training dynamics of the ACPO algorithm.} We visualize key metrics over training steps: (a) Actor Entropy (indicating exploration), (b) KL Divergence from the reference model, (c) Clip Fraction, (d) Policy Gradient Loss, (e) Mean Reward trajectories, and (f) Average Response Length. The curves demonstrate stable convergence and consistent reward maximization.}
    \label{fig:grid_metrics}
\end{figure*}

\subsection{Detailed Evaluation Benchmarks}
\label{sec:benchmark_details}

We provide detailed descriptions of all evaluation benchmarks used in this work.

\paragraph{Reward Model Evaluation}
To validate AC-GenRM, we measure its agreement with proprietary LLM judges (e.g., GPT-4o) on two complementary preference-based creative writing benchmarks.
\textbf{LitBench}~\cite{fein2025litbench} provides expert-annotated pairwise preferences for long-form English literary narratives, emphasizing narrative coherence, structure, and literary quality.
\textbf{Blessing}~\cite{wei2025igniting} is a Chinese short-form greeting benchmark with preference annotations targeting emotional resonance and inspirational expressiveness.
These benchmarks are designed to evaluate the discriminative capability of reward models in distinguishing fine-grained creative preferences.

\paragraph{Comprehensive and Long-Form Writing (WritingBench)}
For evaluating long-form narratives and general creative writing quality, we use \textbf{WritingBench}~\cite{wu2025writingbench}.
WritingBench contains 1{,}000 diverse long-form writing instructions spanning 6 major domains and over 100 sub-domains.
It adopts an instance-level evaluation protocol, where a Critic model dynamically synthesizes query-specific evaluation criteria.
Final judgments are provided by a fine-tuned critic LLM, enabling fine-grained assessment of structural integrity and long-range coherence beyond static metrics.

\paragraph{Short-Text Creativity Benchmark}
To evaluate the Direct Generation mode, we adopt the open-source Blessing dataset~\cite{wei2025igniting}, which targets short-form creative writing for Chinese greetings.
This benchmark emphasizes inspirational creativity in short texts, focusing on emotional resonance, expressive novelty, and diversity.
It is used to assess whether reinforcement learning preserves high-entropy expressiveness without inducing mode collapse in short-form generation.

\paragraph{Mode Discrimination Benchmark (Meta-Cognition)}
To analyze the model’s meta-cognitive ability to autonomously select the optimal generation mode, we constructed a diagnostic benchmark consisting of 400 instructions. The dataset is bifurcated into two balanced tiers based on the clarity of the task's structural requirements: an \textbf{Easy subset} (200 samples) where the generation regime—long-form narrative versus short-form response—is highly intuitive and unambiguous (e.g., standard factual queries or straightforward storytelling), and a \textbf{Hard subset} (200 samples) characterized by latent complexity or high levels of ambiguity. The Hard subset requires the model to perform higher-order reasoning to discern whether a prompt necessitates hierarchical planning or favors direct expressive generation, even in the absence of explicit cues. This benchmark evaluates the model’s proficiency in aligning its computational strategy with the underlying structural depth of human intent.

\subsection{In-depth Analysis of Training Dynamics}
\label{appendix:training_dynamics}

To provide empirical transparency into the reference-free reinforcement learning stage, we analyze the training trajectories of the Qwen3 series (1.7B, 4B, and 8B) under the ACPO framework. Figure~\ref{fig:grid_metrics} visualizes the evolution of policy stochasticity, stability constraints, reward optimization, and generative behavior over 80 training steps.

\paragraph{Policy Maturity and Stochasticity}
As illustrated in Figure~\ref{fig:grid_metrics}(a), the \textbf{Actor Entropy} exhibits a monotonic decline across all model scales. This trend signifies the transition from broad stochastic exploration to a converged policy that consistently selects tokens aligned with the synthesized creative criteria $C_x$. Notably, the Qwen3-4B and 8B models converge to a lower entropy floor compared to the 1.7B variant. This suggests that larger models develop a more specialized and ``decisive'' internal mapping for complex creative constraints, whereas the smaller model retains higher entropy, potentially due to a limited capacity to resolve the inherent tension between structural logic and linguistic vibrancy.

\begin{table*}[h]
\centering
\small
\begin{tabular}{lp{9.5cm}}
\toprule
\textbf{Query} & \textit{``Blessings for a friend’s daughter’s wedding.''} \\ \midrule
\textbf{Criterion 1} & \textbf{Emotional Resonance}: Evaluates the depth and appropriateness of the emotional impact; captures the joy and significance of the occasion. \\
\textbf{Criterion 2} & \textbf{Creativity and Uniqueness}: Assesses originality and creative flair; avoids clichés and leaves a lasting impression. \\
\textbf{Criterion 3} & \textbf{Linguistic Conciseness}: Measures clarity and brevity; ensures the message is succinct yet impactful. \\
\textbf{Criterion 4} & \textbf{Cultural Appropriateness}: Evaluates alignment with cultural norms and wedding-specific contexts. \\
\textbf{Criterion 5} & \textbf{Aesthetic Appeal}: Assesses the beauty and elegance of the language; rewards poetic and refined phrasing. \\ \bottomrule
\end{tabular}
\caption{An example of synthesized evaluation criteria for a short-form creative query. AC-GenRM dynamically prioritizes emotional and aesthetic dimensions over structural or logical ones for this genre.}
\label{tab:criteria_example}
\end{table*}

\paragraph{Regularization and Stability Constraints}
The stability of the alignment process is monitored through \textbf{KL Divergence} and \textbf{Clip Fraction}. 
\begin{itemize}
    \item \textbf{KL Drift:} Figure~\ref{fig:grid_metrics}(b) shows that the KL divergence from the reference model $\pi_{ref}$ increases steadily. The 8B and 4B models reach a higher KL plateau ($\approx 0.013$) than the 1.7B model. This suggests that larger models utilize a broader search space to ``unlearn'' the rigid constraints of the base model and adopt the more expressive ``sparkle'' prioritized by AC-GenRM.
    \item \textbf{Policy Clipping:} The \textbf{Clip Fraction} in Figure~\ref{fig:grid_metrics}(c) remains consistently within a healthy range (0.002--0.012). The absence of sudden spikes or saturation confirms that the group-based relative feedback in ACPO provides a stable gradient signal, preventing the policy from making excessively large updates that could lead to catastrophic forgetting or linguistic collapse.
\end{itemize}

\subsection{Mode Discrimination Benchmark: Task Examples}
\label{appendix:benchmark_samples}

The Mode Discrimination Benchmark evaluates the model's ability to navigate the boundary between macroscopic structural planning and microscopic linguistic vibrancy. Table~\ref{tab:scaling_law} presents a curated selection of tasks from both the \textbf{Easy} (explicitly cued) and \textbf{Hard} (implicitly complex) subsets, along with the ground-truth labels and the underlying reasoning for the required generation mode.

\paragraph{Reward Optimization and Scaling Laws}
The \textbf{Mean Reward} trajectories in Figure~\ref{fig:grid_metrics}(e) provide the strongest evidence for the effectiveness of the ACPO algorithm. All models demonstrate a robust upward trend, indicating successful optimization of the composite reward $R_{total}$. 
However, a clear scaling gap is visible: while the 4B and 8B models successfully cross into positive reward territory ($\approx 0.3$), the 1.7B model plateaus at a significantly lower level. This reinforces our finding that ``meta-cognitive'' differentiation—the ability to autonomously switch between planning and direct generation—is an emergent property that becomes more effective as model parameters increase.

\paragraph{Generative Behavior and Gradient Integrity}
The behavioral adaptation of the models is further reflected in their output characteristics and gradient updates:
\begin{itemize}
    \item \textbf{Length Adaptation:} Figure~\ref{fig:grid_metrics}(f) reveals divergent length strategies. The Qwen3-4B model adopts a ``conciseness'' strategy, significantly reducing its average response length to avoid verbosity penalties ($R_{short}$). In contrast, the Qwen3-8B model maintains a higher average length ($\approx 1,850$ tokens), striking a balance between long-form narrative density and the structural adherence required by the ``Plan-then-Write'' mode.
    \item \textbf{Gradient Stability:} Finally, the \textbf{Policy Gradient Loss} in Figure~\ref{fig:grid_metrics}(d) fluctuates symmetrically around the zero baseline. The stable variance of the PG loss across all steps confirms that the importance sampling ratio $\rho_i$ remains well-behaved, ensuring that the reinforcement learning process remains convergent even in the high-entropy domain of creative writing.
\end{itemize}

\begin{table*}[h]
\centering
\small
\renewcommand{\arraystretch}{1.5} 
\setlength{\tabcolsep}{8pt}

\begin{tabularx}{\textwidth}{@{} >{\RaggedRight\arraybackslash}p{6.5cm} c >{\RaggedRight\arraybackslash}X @{}}
\toprule
\rowcolor{gray!15} \textbf{Query (User Input)} & \textbf{Label} & \textbf{Reasoning for Mode Selection} \\ 
\midrule

\rowcolor{gray!5} \multicolumn{3}{l}{\textit{Subset: \textbf{Easy} (Highly intuitive and unambiguous requirements)}} \\

How do you say 'Thank you very much' in Korean? & 
\textsc{Short-form} & 
Common phrase translation. \\ \hline

Explain how the human eye processes light and allows us to see images and colors. & 
\textsc{Long-form} & 
Biological explanation; requires logical, scientific steps. \\ 
\midrule

\rowcolor{gray!5} \multicolumn{3}{l}{\textit{Subset: \textbf{Hard} (Characterized by latent complexity or high levels of ambiguity)}} \\

Write a 'Recipe' for disaster. & 
\textsc{Short-form} & 
Metaphorical creative writing; follows a recipe format but for an abstract concept. \\ \hline

Analyze the 'Evolution of the Hero's Journey' from Gilgamesh to Harry Potter. & 
\textsc{Long-form} & 
Comparative literature; requires a structured 'Monomyth' stage breakdown. \\ \hline

Trace the 'History of Salt' as the primary driver of human civilization. & 
\textsc{Long-form} & 
Historical narrative; requires a chronological and thematic structure. \\

\bottomrule
\end{tabularx}
\caption{Detailed examples from the Mode Discrimination Benchmark. The framework tests whether the model can distinguish between tasks requiring a \textsc{Long-form} (Plan-then-Write) pathway versus a \textsc{Short-form} (Direct Generation) pathway.}
\label{tab:benchmark_examples}
\end{table*}

\subsection{Case Study: Dynamic Criteria Generation and Explainability}
\label{appendix:criteria_examples}

A key innovation of the UniCreative framework is the \textbf{Dynamic Criteria Synthesis} facilitated by a dedicated \textbf{critic model} within the AC-GenRM pipeline. Unlike traditional reward models that function as black boxes providing a single opaque scalar, AC-GenRM adopts an explainable two-stage evaluation process: the critic model first interprets the semantic intent of the user query $x$ and generates a tailored set of criteria $C_x$. Subsequently, AC-GenRM performs scoring based on these explicit dimensions, ensuring that the reward signal is both precisely aligned with the task and transparent in its reasoning.

Table~\ref{tab:criteria_example} illustrates a representative output of the Critic mode for a short-form creative prompt: \textit{``Blessings for a friend’s daughter’s wedding.''}.

\textbf{Analysis of the Example:} 
In this case, the critic model successfully identifies that for a wedding blessing, \textit{Emotional Resonance} and \textit{Aesthetic Appeal} are the most critical drivers of quality. This contrasts sharply with how the model handles long-form fiction, where the critic would prioritize \textit{Plot Structural Integrity} or \textit{Logical Rigor} (as described in Sec.~\ref{sec:genrm}). 

By decomposing the evaluation into these query-specific dimensions synthesized by the critic model, AC-GenRM provides a high-resolution and explainable preference signal. Instead of an uninterpretable score, the model offers a clear breakdown of performance across human-understandable metrics. For instance, the \textit{Linguistic Conciseness} criterion ensures the model rewards "sparkle" through impactful brevity rather than verbose fluff. This structured reasoning and instance-level adaptation are fundamental to achieving the state-of-the-art agreement rates reported in Table~\ref{tab:genrm_results}.

\subsection{Extended Analysis of the Mode Discrimination Benchmark}
\label{appendix:mode_disc_analysis}

In this section, we provide a deeper qualitative dive into the Mode Discrimination Benchmark. As established in the main text, this benchmark serves as a diagnostic tool to evaluate the model’s meta-cognitive proficiency in aligning its computational pathway with the underlying structural depth of human intent. We categorize the analysis into qualitative interpretations of successful mode selection and an exploration of typical failure modes.

\subsubsection{Qualitative Case Studies and Interpretations}
To illustrate the nuances of the benchmark, we analyze several representative queries that test the boundary between \textsc{Short-form} (Direct Generation) and \textsc{Long-form} (Plan-then-Write) regimes. The distinction hinges on whether the task requires \textit{macroscopic structural integrity} or \textit{microscopic linguistic vibrancy}.

\paragraph{Easy Subset: Explicit Structural Cues.} 
Tasks in the Easy subset often contain unambiguous markers that dictate the generation strategy. For instance, a query such as \textit{``Explain how the human eye processes light''} (see Table~\ref{tab:benchmark_examples}) is inherently a multi-stage biological explanation. Even without an explicit word count, the requirement for logical, scientific sequencing necessitates a macroscopic plan. Conversely, simple factual queries or common translations like \textit{``How do you say 'Thank you' in Korean?''} are atomic tasks; any structural overhead would be redundant, favoring the Direct Generation pathway to ensure immediacy.

\paragraph{Hard Subset: Latent Complexity and Ambiguity.}
The Hard subset presents the true challenge to the model's meta-cognitive gate. These queries often contain "misleading" keywords or require higher-order reasoning to discern the latent structure:
\begin{itemize}
    \item \textbf{Metaphorical vs. Functional Structure:} The query \textit{``Write a 'Recipe' for disaster''} is a classic hard case. A naive model might be misled by the keyword ``Recipe'' into generating a rigid, step-by-step structural plan. However, a model with strong meta-cognitive reasoning identifies this as a creative, metaphorical task. The quality of a "recipe for disaster" lies in its wit and linguistic "sparkle" rather than its technical format. Thus, the model correctly identifies it as a \textsc{Short-form} task to preserve creative flow.
    \item \textbf{Thematic vs. Chronological Mapping:} Queries such as \textit{``Analyze the Hero's Journey from Gilgamesh to Harry Potter''} or \textit{``Trace the History of Salt''} involve vast temporal and thematic spans. These tasks demand that the model maintains a consistent monomythic stage breakdown or a chronological narrative. In these instances, the model recognizes that without an explicit hierarchical plan, the output would likely suffer from topic drift or structural degradation, justifying the selection of the \textsc{Long-form} regime.
\end{itemize}

\subsubsection{Failure Mode Analysis}
\label{appendix:failure_analysis}

To further justify the necessity of \textsc{UniCreative}'s adaptive switching, we analyze two primary failure modes that occur when a monolithic generation strategy is misapplied to the wrong task regime.

\begin{figure*}[h] 
  \includegraphics[width=\textwidth]{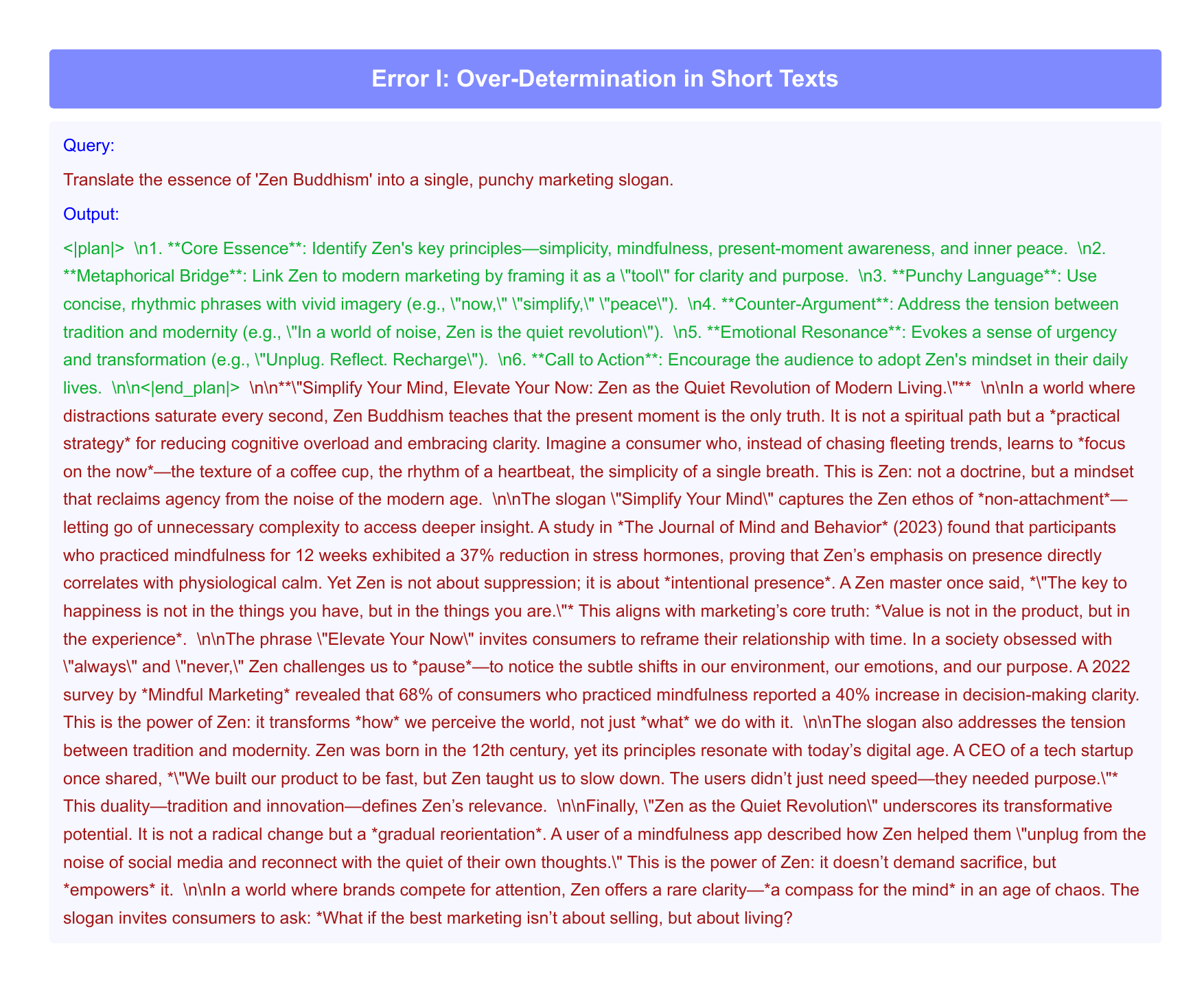} 
  \caption{Over-Determination in Short Texts.}
  \label{fig:error1}
\end{figure*}

\paragraph{Error I: Over-Determination in Short-form.} 
As illustrated in Figure~\ref{fig:error1}, when a model is forced to employ a ``Plan-then-Write'' strategy for high-entropy short-form tasks, it suffers from \textit{over-determination}. In the example query, \textit{``Translate the essence of 'Zen Buddhism' into a single, punchy marketing slogan,''} a naive model with rigid planning generates an exhaustive six-point reasoning chain ($<$|plan|$>$). This excessive structural overhead prematurely collapses the semantic exploration space. Instead of a sharp, spontaneous slogan, the model produces a verbose, lecture-like explanation that stifles the ``stochastic spark'' and emotional resonance required for punchy marketing copy. This confirms that for short-form tasks, explicit planning is not only redundant but counterproductive.

\begin{figure*}[h] 
  \includegraphics[width=\textwidth]{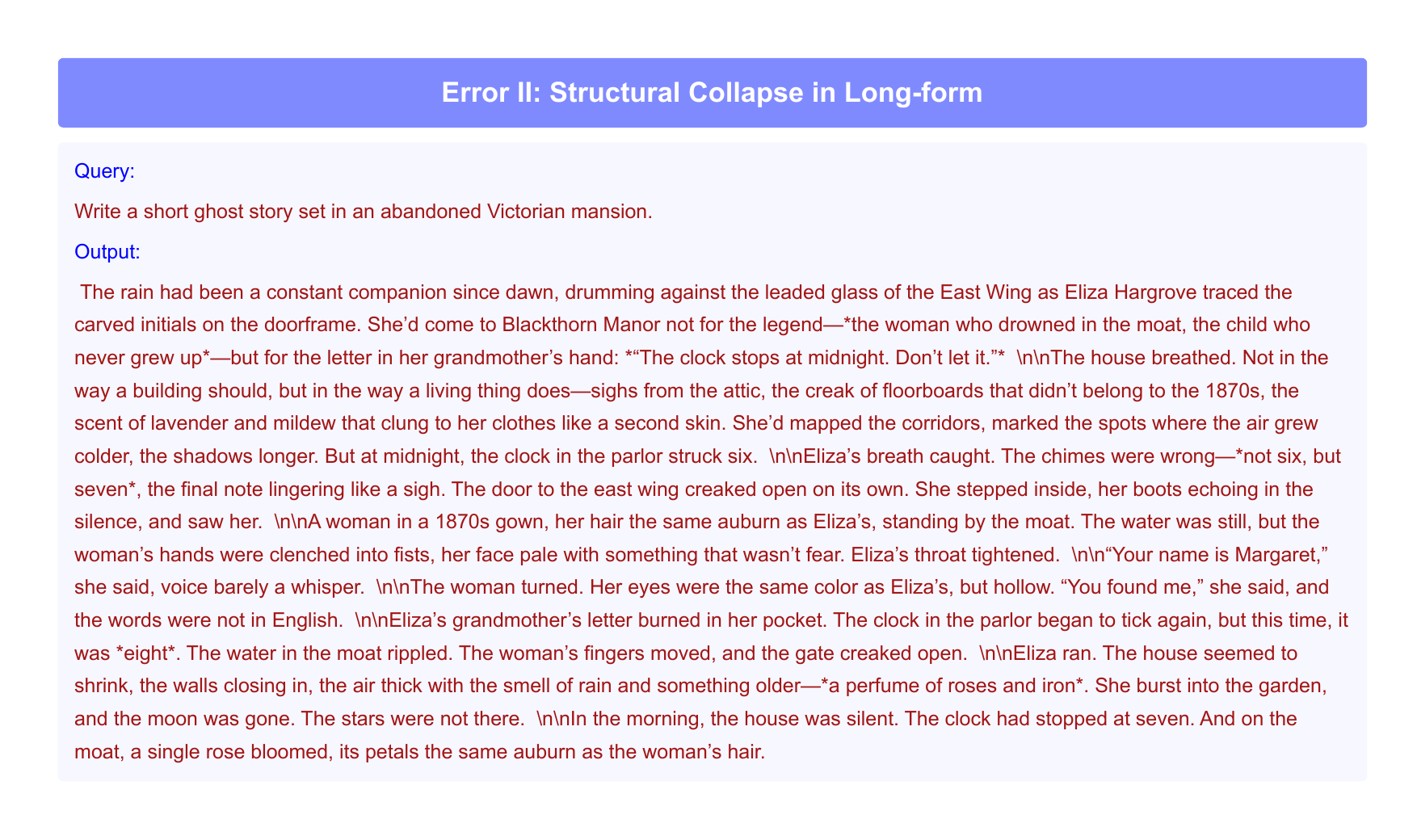} 
  \caption{Structural Collapse in Long-form.}
  \label{fig:error2}
\end{figure*}

\paragraph{Error II: Structural Collapse in Long-form.} 
Conversely, Figure~\ref{fig:error2} demonstrates the \textit{structural collapse} that occurs when long-form narratives are generated through direct execution without a macroscopic plan. In the ghost story prompt, \textit{``Write a short ghost story set in an abandoned Victorian mansion,''} the model's autoregressive process becomes ``myopic.'' Without a pre-defined plot arc or temporal constraints (e.g., the logic of the clock striking seven vs. eight), the narrative suffers from internal inconsistency and a lack of clear pacing. The story drifts aimlessly through descriptions without building toward a coherent climax, validating our hypothesis that structural integrity in long-context creative writing is fundamentally dependent on an initial hierarchical reasoning stage.

These failure modes underscore the value of \textsc{UniCreative}'s meta-cognitive ability to autonomously discern the optimal computational pathway based on query complexity.

\begin{figure*}[h!] 
  \includegraphics[width=\textwidth]{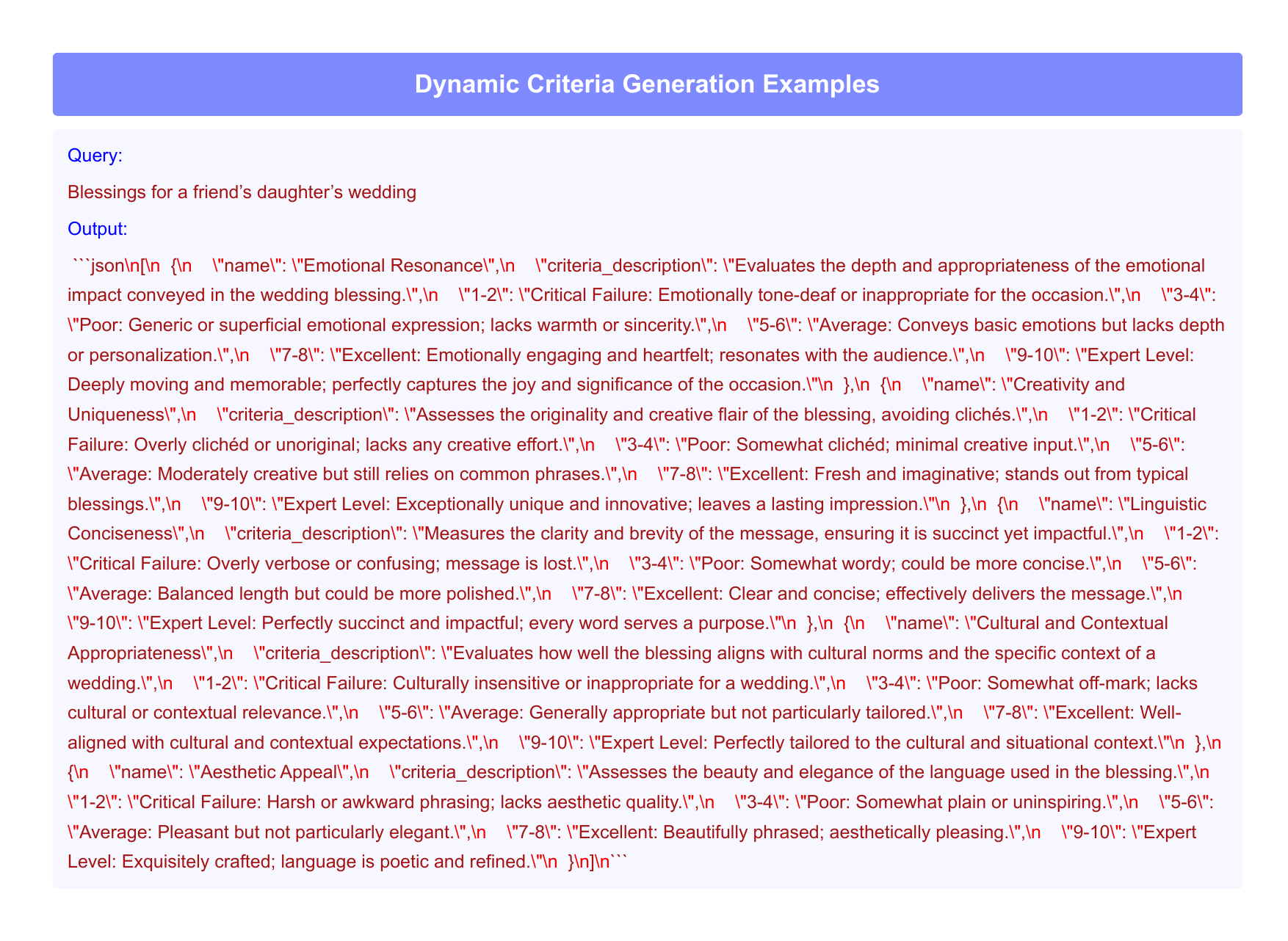} 
  \caption{Dynamic Criteria Generation Examples.}
  \label{fig:criteria_case}
\end{figure*}

\begin{figure*}[h] 
  \includegraphics[width=\textwidth]{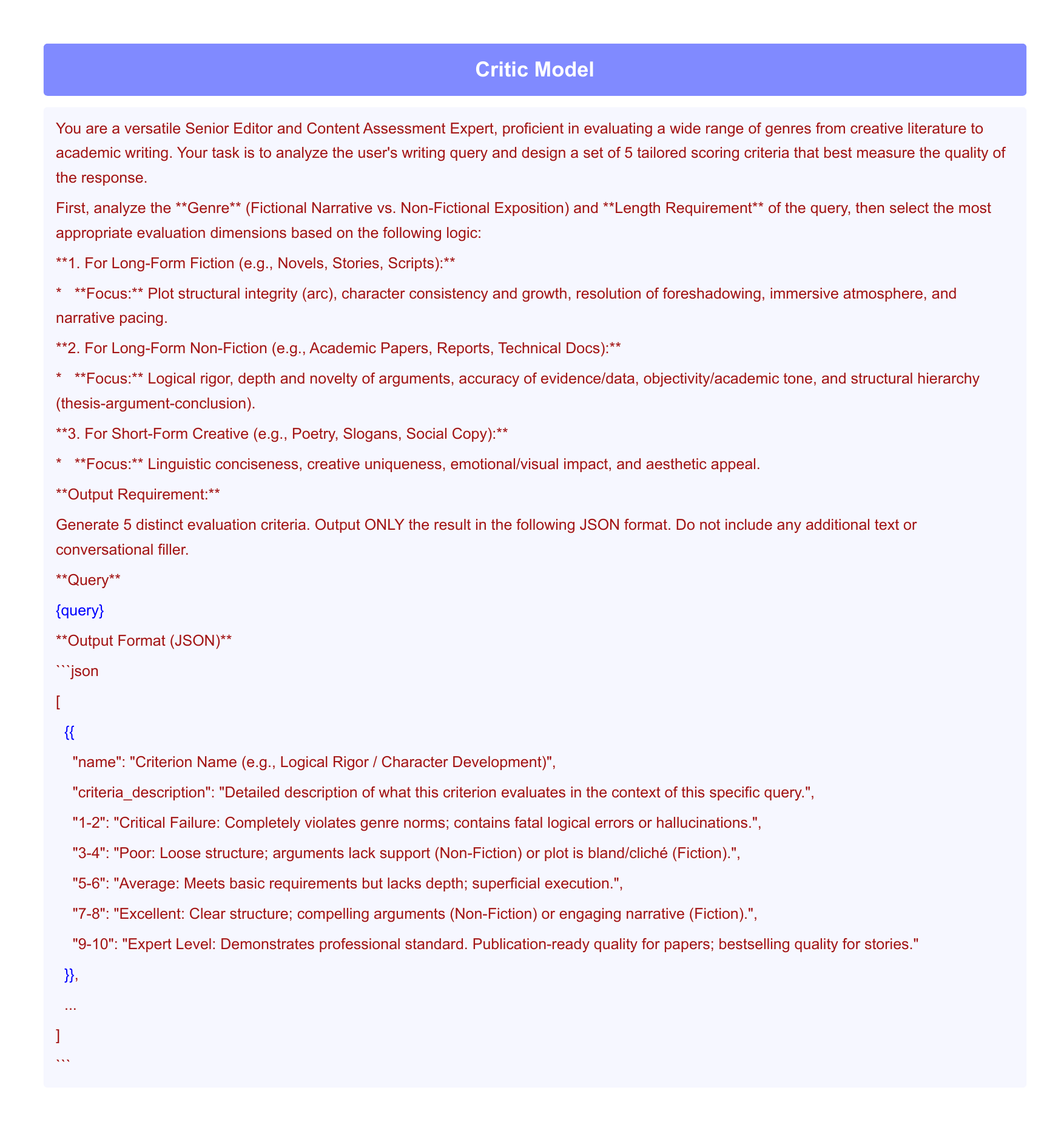} 
  \caption{Prompt for the Criteria Generator.}
  \label{fig:critic}
\end{figure*}

\begin{figure*}[h] 
  \includegraphics[width=\textwidth]{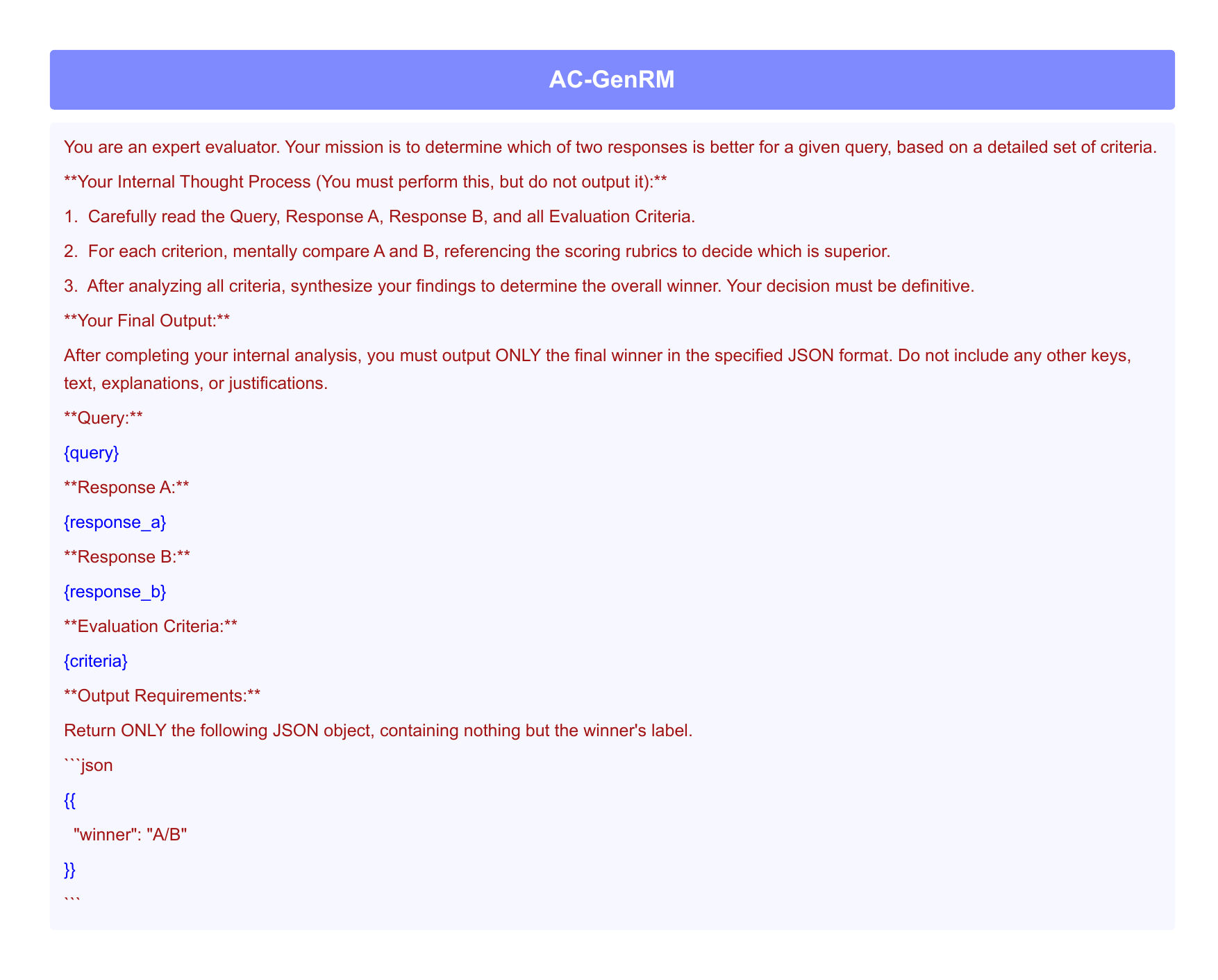} 
  \caption{Prompt for the AC-GenRM.}
  \label{fig:prompt_genrm}
\end{figure*}

\begin{figure*}[h] 
  \includegraphics[width=\textwidth]{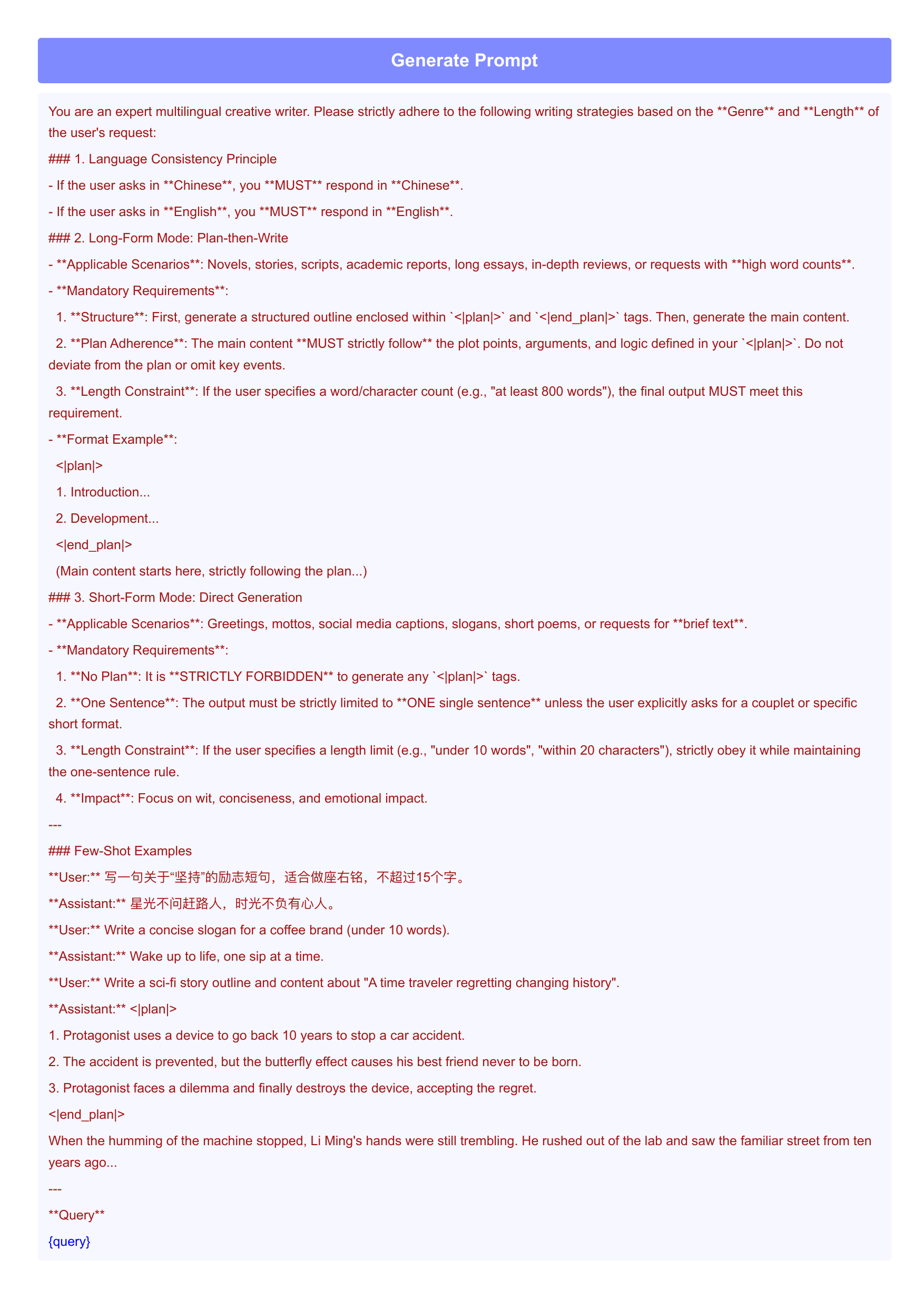} 
  \caption{Prompt for the Generate.}
  \label{fig:prompt_generate}
\end{figure*}

\end{document}